\documentclass{article}
\usepackage{float}
\PassOptionsToPackage{numbers, compress}{natbib}
\usepackage[preprint]{neurips_2026}

\usepackage[utf8]{inputenc}
\usepackage[T1]{fontenc}
\usepackage{graphicx}
\usepackage{amsmath,amssymb,amsthm,mathtools}
\usepackage{hyperref}
\usepackage{url}
\usepackage{xcolor}
\usepackage{booktabs}
\usepackage{subcaption}
\usepackage{placeins}
\usepackage{float}
\usepackage{microtype}
\usepackage{xspace}
\usepackage{enumitem}
\usepackage{array}
\usepackage[normalem]{ulem}

\hypersetup{colorlinks=true,citecolor=blue,linkcolor=blue,urlcolor=blue}

\newcommand{\WW}{\texttt{WeightWatcher}\xspace}

\newcommand{\overlapl}[1]{\mathcal{O}_{l}\!\left(#1\right)}

\newcommand{\MP}{\mathrm{MP}}

\newcommand{\MLP}{{MLP}}
\newcommand{\MA}{{MA}}
\newcommand{\GPT}{{GPT2}}

\newcommand{\safeincludegraphics}[2][]{%
  \IfFileExists{#2}{\includegraphics[#1]{#2}}{%
  \fbox{\begin{minipage}[c][0.10\textheight][c]{0.92\linewidth}\centering
  {\small Missing figure file}\\[2pt]{\small\texttt{\detokenize{#2}}}
  \end{minipage}}}}

\theoremstyle{plain}

\theoremstyle{definition}

\theoremstyle{remark}
\newtheorem*{remark*}{Remark}

\usepackage{xcolor}

\title{Detecting overfitting in Neural Networks during long-horizon grokking using Random Matrix Theory}
\author{
Hari K. Prakash \\
University of California San Diego \\
Data Science and Engineering \\
\texttt{hprakash@ucsd.edu}
\And
Charles H. Martin \\
Calculation Consulting \\
San Francisco, CA, USA \\
\texttt{charles@calculationconsulting.com}
}
\begin{document}
\maketitle 
\begin{abstract}
Training Neural Networks (NNs) without overfitting is difficult; detecting that overfitting is difficult as well. We present a novel Random Matrix Theory method that detects the onset of overfitting in deep learning models without access to train or test data. For each model layer, we randomize each weight matrix element-wise, $\mathbf{W} \rightarrow \mathbf{W}^{\mathrm{rand}}$, fit the shuffled matrix’s empirical spectral distribution with a Marchenko-Pastur distribution, and identify large outliers that violate self-averaging. We call these outliers Correlation Traps. During the onset of overfitting, which we call the "anti-grokking” phase in long-horizon grokking, Correlation Traps form and grow in number and scale as test accuracy decreases while train accuracy remains high. Traps may be benign or may harm generalization; we provide an empirical approach to distinguish between them by passing random data through the trained model and evaluating the JS divergence of output logits.  Our findings show that anti-grokking is an additional grokking phase with high train accuracy and decreasing test accuracy, structurally distinct from pre-grokking through its Correlation Traps. More broadly, we find that some foundation-scale LLMs exhibit the same Correlation Traps, indicating potentially harmful overfitting.
\end{abstract}

\section{Introduction}
Open-weight models are increasingly used as foundations for downstream systems, but it is often unclear whether a checkpoint is robustly trained or overfit to its training distribution. Users may have access to the weights and a model card, but not to the training data, held-out losses, optimizer state, or long-horizon checkpoint history. As a result, two models can look similar from the outside while differing sharply in whether their weights encode useful structure or brittle, data-specific correlations. This motivates a basic question: can we detect signatures of overfitting directly from the weights of a trained model, without having access to the training or any test data?

\begin{figure}[t]
\centering

\begin{minipage}{0.30\columnwidth}
    \centering
  \includegraphics[width=\linewidth]{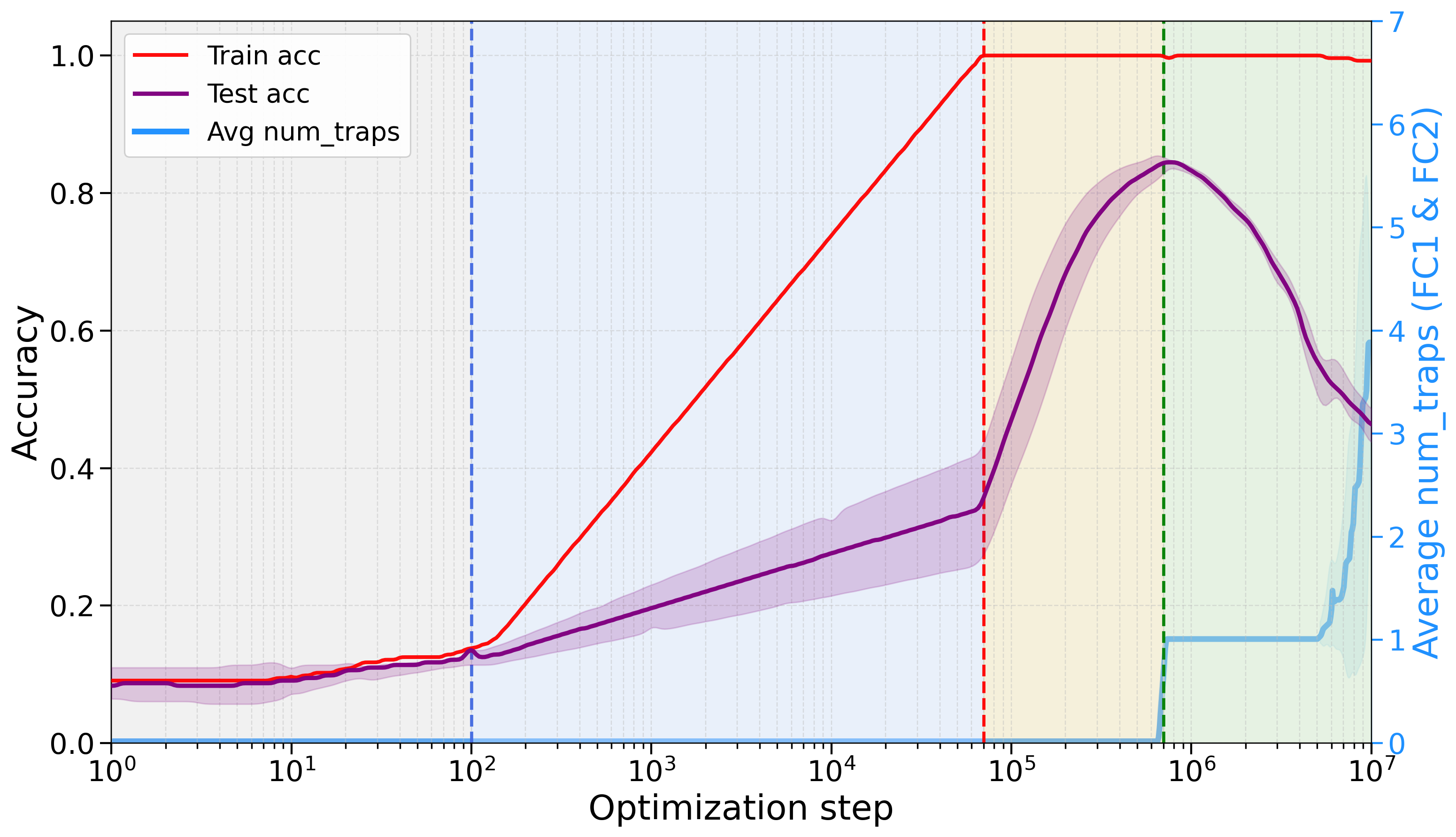}
    \\
    \small (a)~\MLP~
\end{minipage}
\hfill
\begin{minipage}{0.33\columnwidth}
    \centering
    \safeincludegraphics[width=\linewidth]{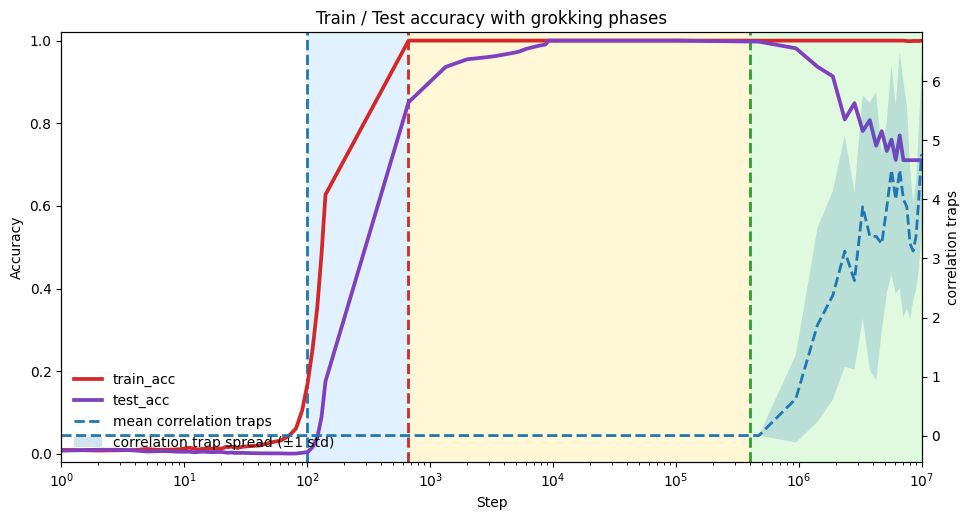}
    \\
    \small (b) Modular Addition (MA) 
    \end{minipage}
\hfill
\begin{minipage}{0.30\columnwidth}
    \centering
    \includegraphics[width=\linewidth]{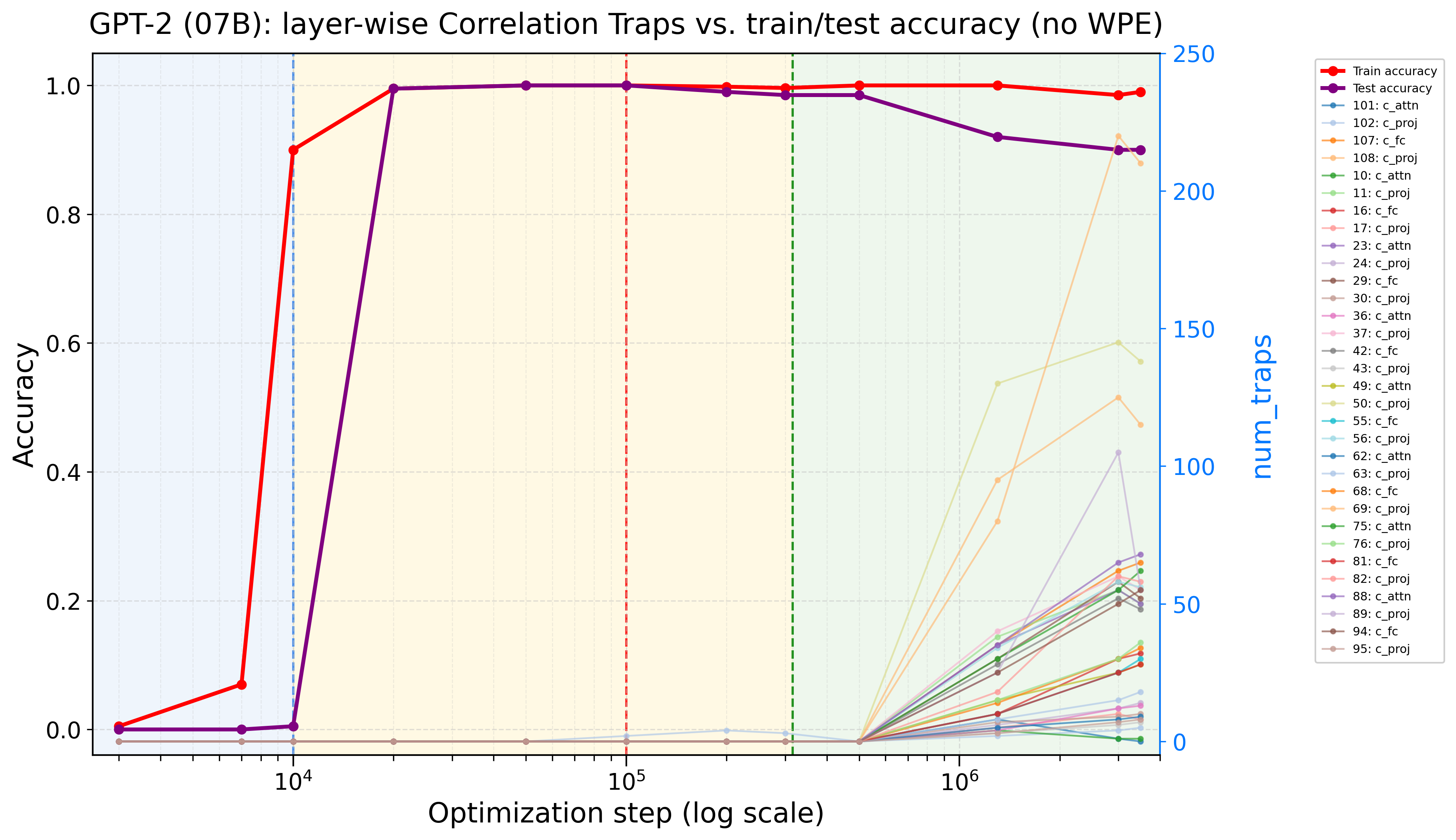}
    \\
    \small (c) GPT2 
\end{minipage}

\caption{\textbf{Grokking dynamics and Correlation Traps.} Reproduction of known grokking experiments, extended to long-horizon training.
  Train accuracy (red), test accuracy (purple), and Correlation Traps (blue) are shown for
  (a) an \MLP~on MNIST, (b) Modular Addition (\MA), and (c) \GPT.  Shaded regions denote pre-grokking
  (gray), grokking (yellow), and anti-grokking (green).  Correlation Traps arise at the onset of late-stage
  test-accuracy decline and track it as it decreases.}

\label{fig:training_curves}
\end{figure}

Because such histories are usually unavailable for open-weight checkpoints, we first study a setting where the relevant dynamics are visible and overfitting can be readily induced with long-horizon training: \emph{grokking}.  In grokking, training accuracy reaches near perfection while test accuracy stays near chance for many optimization steps, before abruptly improving~\citep{power2022grokking}. Grokking has been studied across several architectures and tasks, including algorithmic tasks such as modular addition, computer vision models, and GPT-style transformers~\citep{nanda2023progress,wang2024grokked}. We extend this view to the long-horizon after grokking, where long-horizon training can drive the model into a classical overfitting phase.  We call this post-generalization  regime \emph{anti-grokking}.

Our setup lets us compare three phases of learning: pre-grokking, grokking, and anti-grokking. Pre-grokking and anti-grokking can look deceptively similar from train and test accuracy alone: in both regimes, training accuracy is high while held-out accuracy is poor. But they are structurally different. Pre-grokking occurs before the model has found a generalizing solution; anti-grokking occurs after such a solution has been found and then lost. In comparing different phases, this makes long-horizon grokking an excellent case for studying harmful overfitting.

To detect this post-grokking overfitting structure, we introduce \emph{Correlation Traps}. Correlation Traps are outliers to the (self-averaging) Random Matrix Theory (RMT) description of the layer weight matrices.
For each layer, we shuffle the weights entry-wise, $\mathbf{W}\rightarrow\mathbf{W}^{\rm rand}$, fit the eigenvalue spectrum of the covariance matrix $\mathbf{X}^{\rm rand}=\frac{1}{N}(\mathbf{W}^{\rm rand})^\top\mathbf{W}^{\rm rand}$ to a Marchenko-Pastur (MP) bulk, and count outliers far beyond the MP edge. The key observation is that while both pre-grokking and anti-grokking can look similar, only anti-grokking exhibits Correlation Traps.

Figure~\ref{fig:training_curves} shows the phenomenon in three controlled settings: an MNIST~\MLP, a transformer trained on Modular Addition (\MA), and a GPT2 transformer trained for long-horizon grokking~\citep{wang2024grokked}. In all three cases, training separates into three phases. During pre-grokking, training accuracy increases while test accuracy remains low. During grokking, test accuracy improves and the model reaches a high-generalization solution. During anti-grokking, test accuracy falls despite sustained near-perfect training accuracy. Across all three tasks, the trap signal tracks this late divergence: trap counts are low while generalization improves and increase as test accuracy declines.

These results indicate that Correlation Traps reveal the onset of this overfitting phase in a well-studied grokking setting where standard train/test curves alone would not distinguish it from pre-grokking.  
\paragraph{Our Contributions.}
\begin{itemize}[leftmargin=1em]
    \item We define Correlation Traps as outliers to the MP/TW right-edge in shuffled layer spectra.
    \item We show the onset of traps tracks anti-grokking in \MLP, \MA, and \GPT-style runs.
    \item We link traps to non-self-averaging through localization and condensation.
    \item We distinguish between harmful and benign traps using  JSD-based behavioral testing.
    \item We screen GPT-OSS 20B/120B checkpoints with layer-wise trap profiles.
\end{itemize}
Most importantly, however, we argue that Correlation Traps can be used to detect signatures of potentially harmful overfitting in open-source, foundation-scale models from just the weights.

\section{Related Work}

\paragraph{Grokking and long-horizon generalization.}
Grokking was introduced by Power et al.~\citep{power2022grokking} as delayed generalization after training accuracy has already saturated. Subsequent work has studied grokking in algorithmic and mechanistic settings, including modular arithmetic, MLPs, and Transformer models~\citep{liu2022towards,nanda2023progress,lee2024grokking,wang2024grokked}. Most of this literature focuses on the transition from memorization to generalization: the model first fits the training set, then later discovers a rule that generalizes. Our focus is the long-horizon regime after this transition, where a model that has already grokked can lose held-out performance  (and overfit its training data) after extended training.

\paragraph{Random-matrix diagnostics of neural-network weights.}
Our diagnostic builds on spectral approaches to neural-network weight matrices, especially Random Matrix Theory (RMT) and the \texttt{WeightWatcher} framework~\citep{weightwatcher,martin2025setol,martin2021implicit,martin2021predicting}. Prior spectral diagnostics use heavy-tailed structure of the correlated (unrandomized) layer weight matrices $\mathbf{W}$, and related metrics to characterize trained networks. We analyze the spectral properties of the randomized $\mathbf{W}$, and look for large eigenvalues $(\lambda_{trap})$ that deviate from the Random Matrix Theory (RMT) Marchenko-Pastur (MP) baseline. We call these outliers  Correlation Traps, and track them through extended training, connecting them to overfitting in the anti-grokking phase.  We note that Correlation Traps were first proposed in \cite{martin2025setol}.

\paragraph{Self-averaging and overfitting.}
Our overfitting criterion connects to statistical-mechanics accounts of glassy learning, where poor generalization reflects sample-specific structure rather than a single stable rule~\citep{Hopfield1982,Amit1985,gardner1988space,seung1992statistical,bos1998weightdecay}. The MP law gives a self-averaging baseline for randomized layer spectra, and Correlation Traps violate that baseline. Such traps can support a non-self-averaging generalization error such as through localization, where a small coordinate set retains $O(1)$ variance under subsampling, or through condensation, where a dominant spectral mode carries macroscopic variance.\footnote{The corresponding physics analogies are Anderson localization, Bose--Einstein condensation, and the Curie--Weiss mean-field model of magnetization~\citep{anderson1958absence,bose1924plancks,einstein1925quantentheorie,weiss1907hypothese}.}In either case, the presence of traps aligns with classic notions of overfitting that occur in the spin-glass-like phase(s) of simple statistical mechanics models of NNs.

\section{\texorpdfstring{\texttt{WeightWatcher}, Random Matrix Theory, and Correlation Traps}{WeightWatcher, Random Matrix Theory, and Correlation Traps}}
\label{sec:ww-rmt}

The open-source \texttt{WeightWatcher} tool (\WW) implements several random-matrix-based analyses of neural-network layers~\cite{weightwatcher}.  Here, 
it is used to examine the individual layer spectral densities using techniques adapted from Random Matrix Theory (RMT), and rigorously established by Statistical Mechanics and extensive experimental observations.

\subsection{The Marchenko-Pastur (MP) self-averaging baseline}

For a layer weight matrix $\mathbf{W}\in\mathbb{R}^{N\times M}$, define the layer covariance
\begin{equation}
\mathbf{X}
:=
\frac{1}{N}\mathbf{W}^{\top}\mathbf{W}
\in \mathbb{R}^{M\times M}.
\label{eq:layer-covariance}
\end{equation}
Let $\{\lambda_i\}_{i=1}^{M}$ be the eigenvalues of $\mathbf{X}$. The empirical spectral density (ESD) is
\begin{equation}
\rho_{\mathrm{emp}}(\lambda)
:=
\frac{1}{M}\sum_{i=1}^{M}\delta(\lambda-\lambda_i).
\label{eq:esd}
\end{equation}
If the entries of $\mathbf{W}$ are i.i.d.\ and well behaved, then in the limit $N,M\to\infty$, with $Q=N/M\ge 1$ fixed, $\rho_{\mathrm{emp}}(\lambda)$ converges to the Marchenko--Pastur density~\cite{marchenko1967distribution}. 
\begin{equation}
\rho_{\MP}(\lambda)
=
\frac{Q}{2\pi\sigma^2}
\frac{\sqrt{(\lambda_+-\lambda)(\lambda-\lambda_-)}}{\lambda}
\mathbf{1}_{[\lambda_-,\lambda_+]}(\lambda),
\label{eq:mp-density}
\end{equation}
with edges
\begin{equation}
\lambda_{\pm}
=
\sigma^2\left(1\pm Q^{-1/2}\right)^2.
\label{eq:mp-edges}
\end{equation}
At  finite $N$, the right edge $\lambda_+$ lives within the scale of the Tracy-Widom (TW) fluctuations.  

Also, and importantly, the MP singular and/or eigenvectors $\mathbf{v}$ are delocalized with randomly distributed components. That is, we may expect $|v_j|^2 \sim \tfrac{1}{M}$
up to fluctuations.

The MP bulk is consistent with \textbf{self-averaging} behavior. In the large-$N$ limit, the ESD concentrates onto the deterministic MP law,
\[
\rho_{\mathrm{emp}}(\lambda)\;\to\;\rho_{\mathrm{MP}}(\lambda),
\quad N,M\to\infty,\; Q=\tfrac{N}{M}\ \text{fixed}.
\]
Large spectral outliers represent \textbf{non-self-averaging} structure: a small number of directions dominate the statistics. Observables that depend on these directions fail to concentrate. This is the crux of our analysis: such outliers $(\lambda_{trap},\mathbf{v}_{trap})$  can be detected directly from the weight matrices and indicate that the layer has developed structured, non-random correlations associated with overfitting.

\begin{figure}[t]
\centering

\begin{minipage}{0.32\columnwidth}
    \centering
    \includegraphics[width=\linewidth]{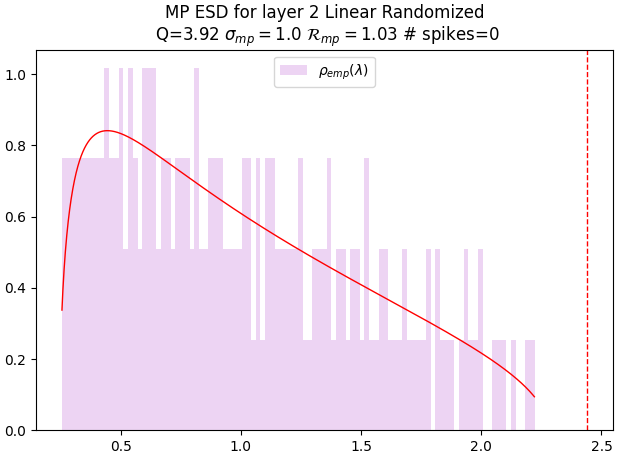}
    \\
    \small (a)
\end{minipage}
\hfill
\begin{minipage}{0.32\columnwidth}
    \centering
    \includegraphics[width=\linewidth]{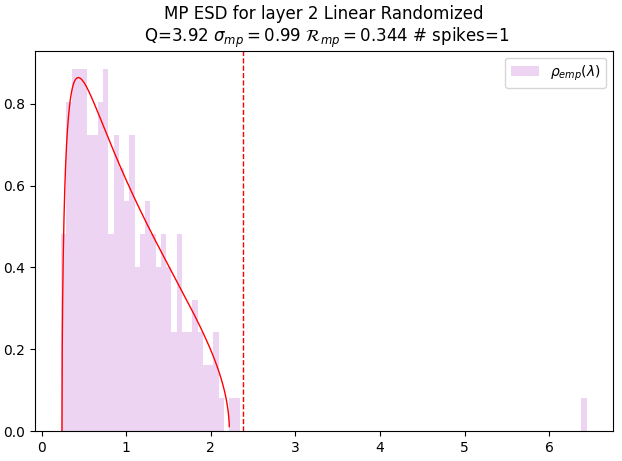}
    \\
    \small (b)
\end{minipage}
\hfill
\begin{minipage}{0.32\columnwidth}
    \centering
    \includegraphics[width=\linewidth]{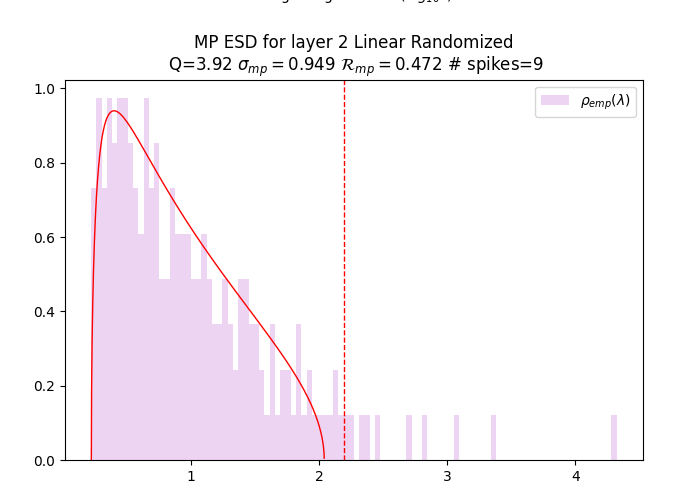}
    \\
    \small (c)
\end{minipage}

\caption{
\textbf{Randomized ESDs, MP fits, and emergence of Correlation Traps.}
(a) Example of the ESD of $\mathbf{X}^{\mathrm{rand}}$ compared with a Marchenko--Pastur (MP) fit. In a self-averaging randomized layer, the spectrum is described by the MP bulk and no large right-edge outliers remain.
(b,c) Examples of trapped randomized spectra from Layer~2 of the~\MLP. Right before collapse, the randomized spectrum already shows a dominant spike near $\lambda_{\mathrm{trap}}\approx10^{6.5}$; during the final anti-grokking stage, multiple spikes lie well beyond the MP bulk. We call these outliers Correlation Traps.
}
\label{fig:ESDs}
\end{figure}

\subsection{Entry-wise shuffling and Correlation Traps}

Recognizing that RMT requires the elements of a matrix to be i.i.d. and well-behaved (i.e., not too heavy tailed), to apply RMT faithfully, we randomize the weight matrix elementwise, and form
\begin{equation}
\mathbf{W}
\mapsto
\mathbf{W}^{\mathrm{rand}}
\label{eq:entrywise-shuffle}.\;\;
\end{equation}
Under the randomized null, the entries of $\mathbf{W}^{\mathrm{rand}}$ should be effectively uncorrelated, so the ESD of $\mathbf{X}^{\mathrm{rand}}=\tfrac{1}{N}(\mathbf{W}^{\mathrm{rand}})^{\top}\mathbf{W}^{\mathrm{rand}}$
should be well fit by an MP distribution if the model is well-trained. Also, the  singular vectors  of $\mathbf{W}^{\mathrm{rand}}$ (and eigenvectors of $\mathbf{X}^{\mathrm{rand}}$) will be delocalized.
\footnote{The bulk MP vector components $v_{i}$ will follow a Porter–Thomas distribution~\cite{porter1956fluctuations, potters2021first}}.

Figure~\ref{fig:ESDs} shows this diagnostic. In a well-behaved layer, the randomized ESD follows the fitted MP bulk and no large right-edge outliers appear; see Fig.~\ref{fig:ESDs}(a).  During anti-grokking, however, the randomized spectrum develops separated spikes well beyond the MP edge; see Fig.~\ref{fig:ESDs}(b,c).  These spikes are structural outliers that survive entry-wise shuffling (and frequently localize).

We call these outliers \textbf{Correlation Traps}. If $\lambda_{trap}$ is an eigenvalue of $\mathbf{X}^{\mathrm{rand}}$ and $\lambda_+^{\MP}$ is the fitted MP right edge, then a trap is an eigenvalue that lies significantly beyond the right MP edge:
\begin{equation}
\text{Correlation Trap:}
\qquad
\lambda_{trap}
>
\lambda_+^{\MP}
+
\Delta_{\mathrm{TW}}
\label{eq:correlation-trap}
\end{equation}
where $\Delta_{\mathrm{TW}}$ denotes the scale of finite-size Tracy-Widom  fluctuations.

By the Baik--Ben Arous--P\'ech\'e (BBP) theory, an eigenvalue that separates beyond the TW edge   is a structural outlier \cite{baik2005phase}. This is why the MP fit provides the randomized null for our diagnostic. Moreover, the resulting outliers are similar in spirit to the strong aligned directions emphasized by Li and Sonthalia \cite{li2026risk} which induce catastrophic failure.  

Our setting is nonparametric and implemented as a practical diagnostic, and does not distinguish the specific mechanisms that may be inducing what appears to be non-self-averaging/failure-of-concentration.  Also, some traps are harmful, other benign.  Later, we show how to distinguish between harmful and benign traps, again without needing any test or training data, in a given model.
\begin{table}[h]
\centering
\caption{\textbf{Trap-detection algorithm.} Full procedure used to compute trap-count curves.}
\label{tab:trap_algorithm}
\vspace{0.15em}
\setlength{\tabcolsep}{6pt}
\renewcommand{\arraystretch}{1.05}
\begin{tabular}{@{}c p{0.84\linewidth}@{}}
\toprule
\textbf{Step} & \textbf{Procedure} \\
\midrule
1 &
Take a trained layer $\mathbf{W} \in \mathbb{R}^{N \times M}$ at a saved checkpoint. \\

2 &
Shuffle the entries $W_{ij}$ elementwise to form $\mathbf{W}^{\mathrm{rand}}$. \\
3 &
Run SVD($\mathbf{W}^{\mathrm{rand}}$), form the eigenvalue density, $\rho_{emp}(\lambda)$ (or ESD)  \\
4 & Fit the ESD $\rho_{emp}(\lambda)$ to the Marchenko--Pastur (MP) law. \\
5 &
Identify eigenvalues $\lambda_{trap} > \lambda_{+}^{\mathrm{MP}} + \Delta_{\mathrm{TW}}$; these are Correlation Traps \\
\bottomrule
\end{tabular}
\end{table}

The \WW tool detects Correlation Traps automatically, as outlined in Table~\ref{tab:trap_algorithm}.

\section{Traps as Signs of Non-Self-Averaging: Localization and Condensation}
\label{sec:trap-mechanisms}

Self-averaging is the statistical-mechanics analogue of concentration.
An observable $A_N$ self-averages when its relative fluctuations vanish,
for example when
\begin{equation}
    \frac{\operatorname{Var}(A_N)}{\mathbb{E}[A_N]^2}
    \longrightarrow 0 .
    \label{eq:self-averaging}
\end{equation}
For the empirical risk $R_n$, one defines the variance and covariance:
\begin{equation}
    R_n = \frac{1}{n}\sum_{i=1}^n L_i ,
    \qquad
    \operatorname{Var}(R_n)
    =
    \frac{1}{n^2}\mathbf{1}^{\top}\Sigma_n\mathbf{1},
    \quad
    (\Sigma_n)_{ij}=\operatorname{Cov}(L_i,L_j).
    \label{eq:risk-variance}
\end{equation}
 Concentration requires that the covariance has no  macroscopic, sample-specific mode(s). More generally, define a covariance-like
matrix $C$ with eigendecomposition
\begin{equation}
    C = \sum_{\alpha} \lambda_{\alpha}
    v_{\alpha}v_{\alpha}^{\top}.
\end{equation}
Let $A_b$ be some linearized observable with sensitivity vector $b$. Write the variance of $A_b$ as
\begin{equation}
    \operatorname{Var}(A_b)
    =
    b^{\top} C b
    =
    \sum_{\alpha}
    \lambda_{\alpha}
    \langle b,v_{\alpha}\rangle^2 .
    \label{eq:mode-variance}
\end{equation}
A trap is therefore dangerous when one term in this sum remains
macroscopic. A \emph{failure to concentrate} can happen in at least two different ways: \emph{localization} and/or \emph{condensation}.

\paragraph{Geometric localization.}
A trap mode may be localized in coordinates. For a normalized eigenvector
$v\in\mathbb{R}^M$, this means that its squared mass is concentrated on a
small set $S$:
\begin{equation}
    \sum_{i\in S} v_i^2 \approx 1,
    \qquad |S|\ll M .
\end{equation}
The corresponding variance component $\lambda vv^{\top}$ is then carried
by a small effective support. If an observable $(A_b)$ has non-negligible projection on
this support, then it is not averaging over many comparable coordinates, so its
fluctuations need not decay at the usual rate.

In practical terms, this means the trap is highly localized, and frequently contains a significant fraction of the top $5\%$ mass of $W_{ij}$.  We examined the localization of traps and found that they provide only limited predictive signal for harmful traps, with the relationship varying across experiments.
\footnote{This is analogous to Anderson localization: an eigenmode is spatially or
coordinate-localized, so the relevant variance is supported on only a few
degrees of freedom.}

\paragraph{Spectral condensation.}
A trap may also be diffuse in coordinates but dominant in spectrum. Suppose
one eigenvalue $\lambda_{\rm max}$ is much larger than the rest.
Then the variance contribution
\begin{equation}
    \lambda_{\rm max} \langle b,v_1\rangle^2
\end{equation}
can dominate Eq.~\eqref{eq:mode-variance}, even when $v_1$ is spread across
many coordinates. In this case self-averaging fails not because the mode is
geometrically sparse (i.e. the trap is highly localized), but because variance has condensed into a single
large spectral degree of freedom.\footnote{This resembles
the Curie-Weiss model, which has a matrix $(J/N)\mathbf{1}\mathbf{1}^{\top}$ with one large
eigenvalue and a perfectly delocalized eigenvector
$\mathbf{1}/\sqrt{N}$. Similarly, a Bose condensate occupies one spectral
mode that can remain spatially diffuse.}

Correlation Traps can exhibit localization, condensation, or a mixture of
both. The MP violation indicates non-self-averaging structure, but it is
not a sign rule: localized traps and diffuse spectral spikes can each be
useful, harmful, or benign. Their effect on generalization must therefore
be determined either by intervention and/or the data-free JSD diagnostic ablation test  described below.

\section{Empirical Results: Correlation Traps Track Anti-Grokking}

We study three standard grokking benchmarks, denoted~\MLP~, \MA, and~\GPT. The first, 
\MLP, is a fully connected depth-3 ReLU MLP trained on a balanced MNIST subset containing 100 examples from each class, for a total of 1{,}000 training points. The network has width $M=200$ in each hidden layer and is trained with AdamW, MSE loss on one-hot targets, and learning rate $\text{lr}=5\times 10^{-4}$ for up to $10^7$ optimization steps. The main runs use \texttt{WD=0}, and we include a control with \texttt{WD=0.01} to measure the effect of weight decay. Full hyperparameters appear in Appendix~\ref{app:exp_setup}.

The second benchmark, \MA, is a small one-layer transformer trained on the modular-addition task $\mathbf{x}+\mathbf{y}\bmod P$. We again train far beyond the point at which the model first generalizes. Architecture details, phase summaries, and per-layer trap counts appear in Appendix~\ref{app:modadd}. Pre-grokking, grokking, and anti-grokking are descriptive phase labels tied to the observed train/test trajectories.


The third benchmark, \GPT, follows the synthetic composition task of GrokkedTransformer - Wang et al. ~\citep{wang2024grokked}. It uses a synthetic knowledge graph with atomic one-hop facts $(h,r,t)$, queried as $(h,r)\mapsto t$, and latent composition rules that produce inferred two-hop facts. For instance, atomic facts $(h,r_1,m)$ and $(m,r_2,t)$ imply the inferred query $(h,r_1,r_2)\mapsto t$. Following the original setup, atomic facts are split into $\mathrm{atomic}{\mathrm{ID}}$ and $\mathrm{atomic}{\mathrm{OOD}}$. Training includes all atomic facts plus a random subset of inferred facts derived only from $\mathrm{atomic}{\mathrm{ID}}$, denoted $\mathrm{train\_inferred}{\mathrm{ID}}$, and a hold-out set, $\mathrm{test\_inferred}_{\mathrm{ID}}$. Thus, ``test’’ measures in-distribution compositional generalization over unseen inferred facts from the same atomic pool. See Appendix~\ref{app:gpt2} for more details.

 The supplemental material (See Appendix~\ref{app:repro}) contains the corresponding training and analysis code.

\subsection{MLP: a third phase after grokking}
\begin{figure}[h]
\centering

\begin{minipage}{0.31\columnwidth}
    \centering
  \includegraphics[width=\linewidth]{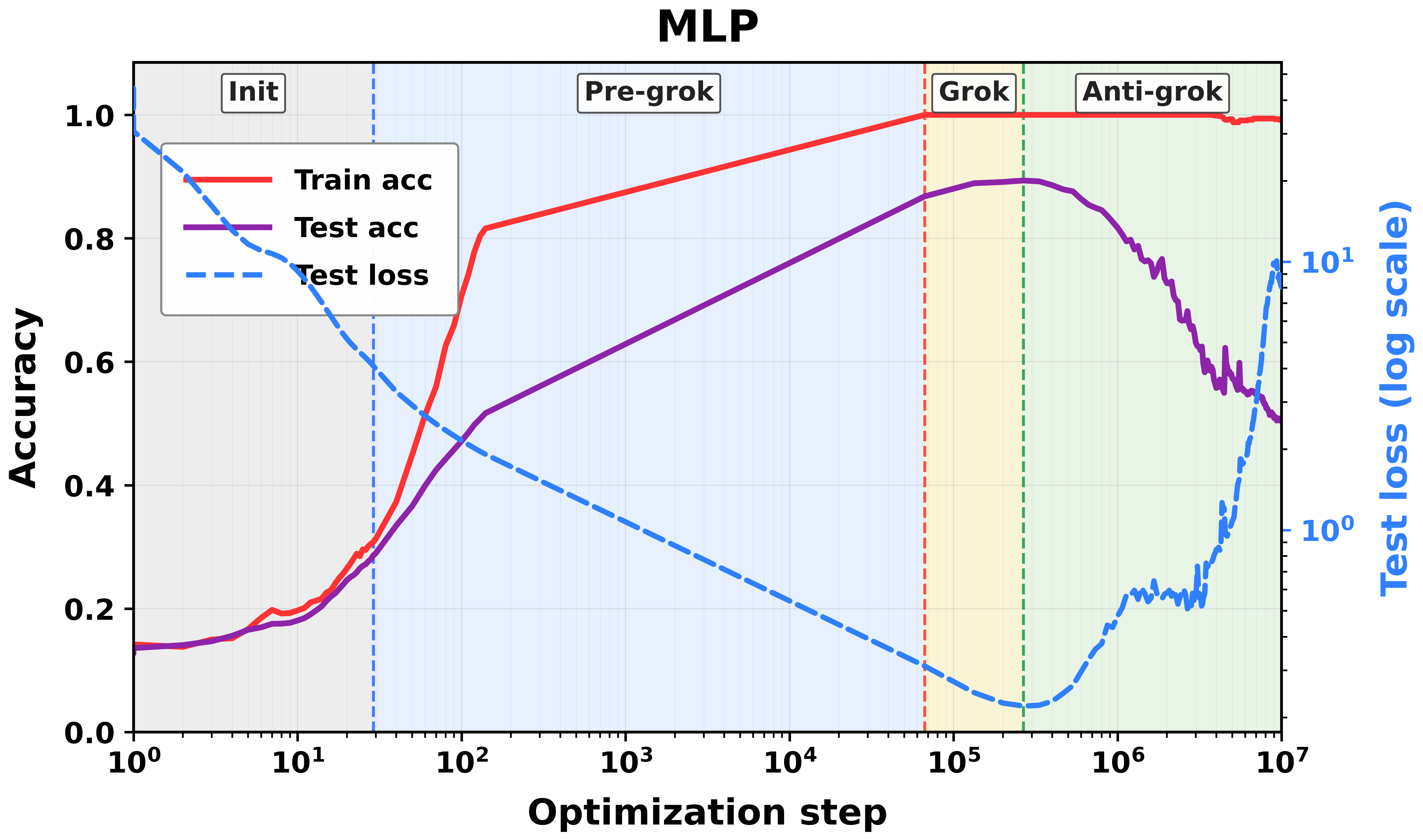}
    \\
    \small (a)~\MLP
\end{minipage}
\begin{minipage}{0.32\columnwidth}
    \centering
    \includegraphics[width=\linewidth]{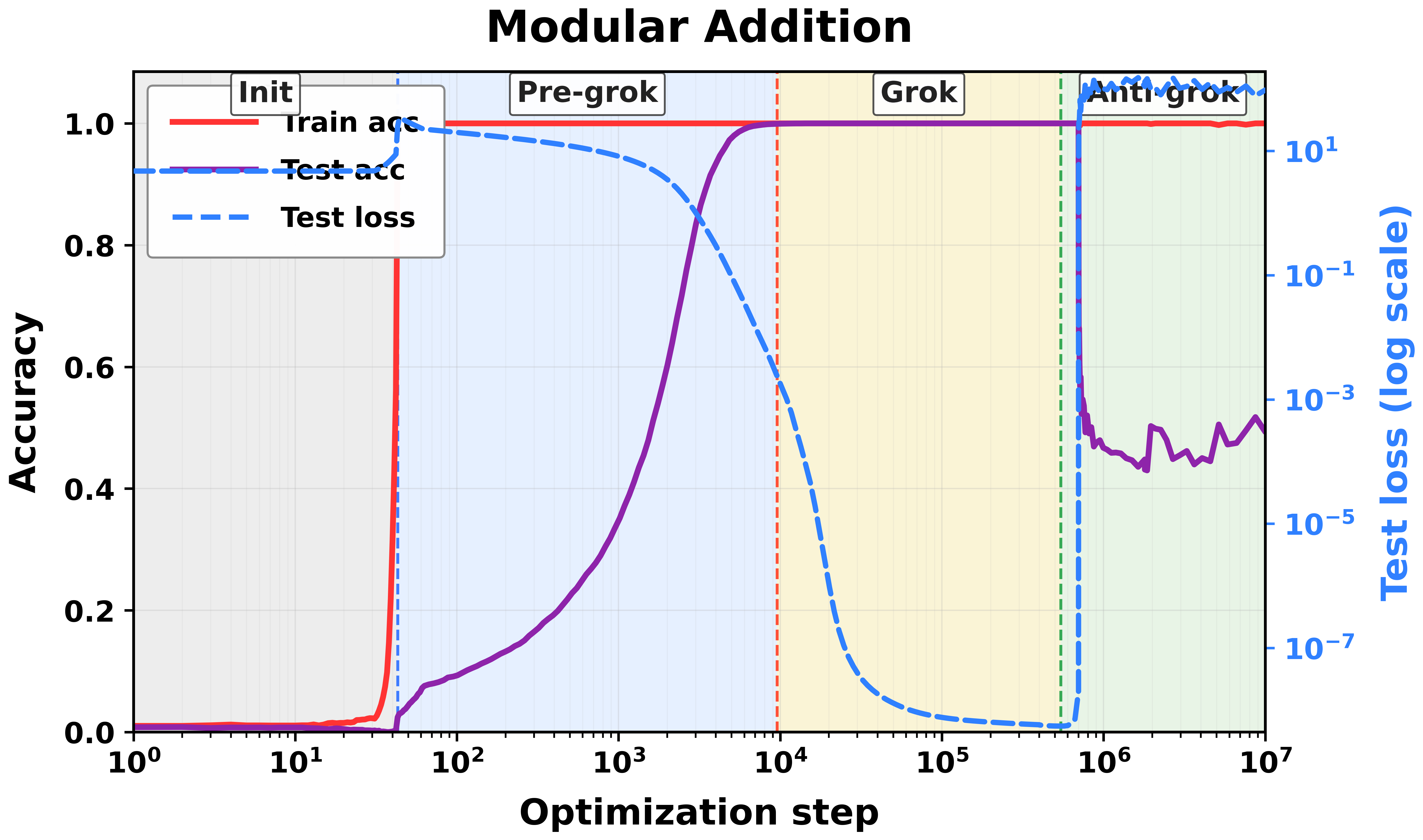}
    \\
    \small (b)~\MA
\end{minipage}
\begin{minipage}{0.33\columnwidth}
    \centering
    \includegraphics[width=\linewidth]{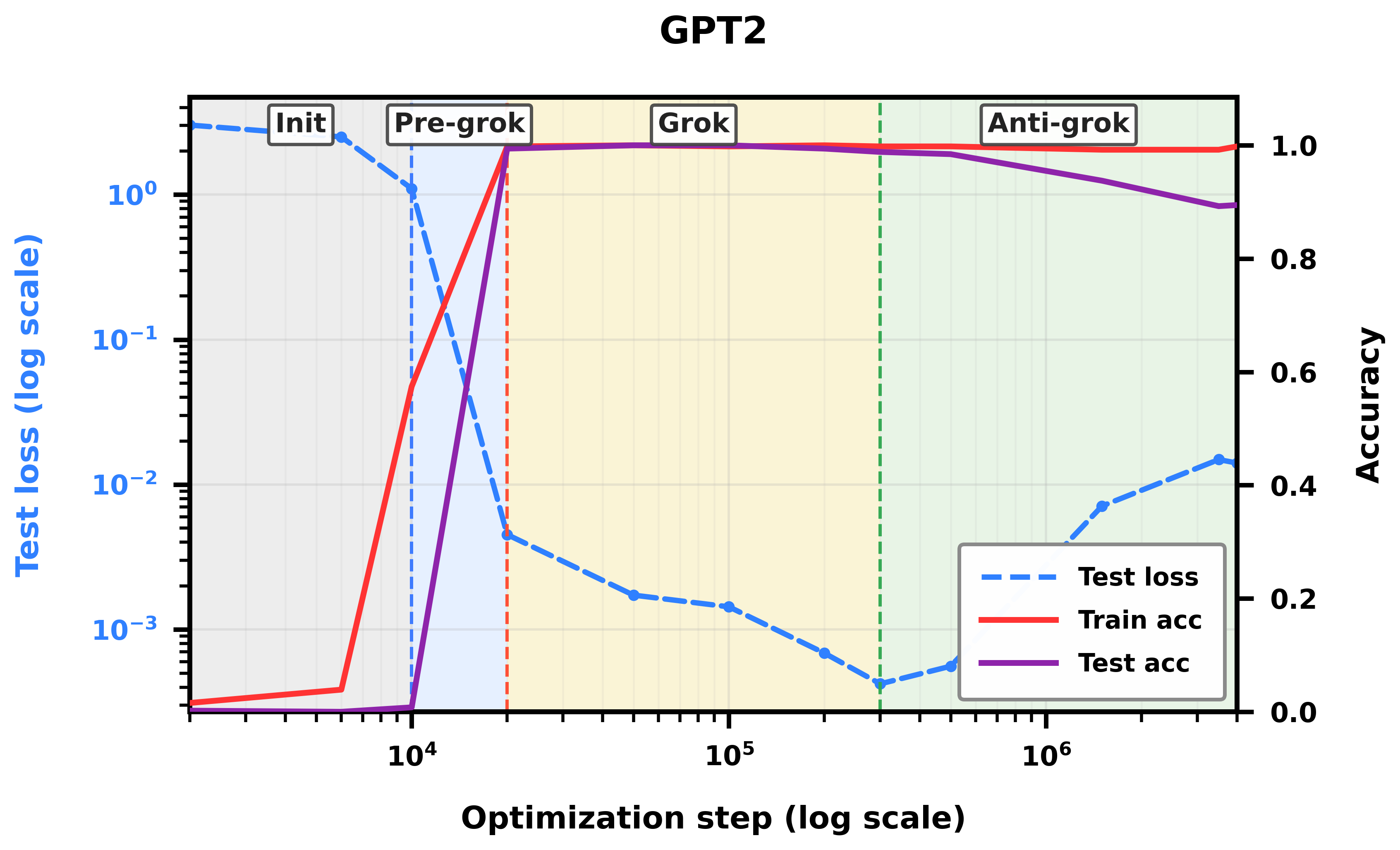}
    \\
    \small (c)~\GPT
\end{minipage}

\caption{\textbf{Grokking dynamics and test loss.}
  Train accuracy (red), test accuracy (purple), and test loss  (blue) are shown for
  (a) an MLP on MNIST and (b) Modular Addition  (\MA) and (c) GPT2.}
\label{fig:training_losses}
\end{figure}

Figure~\ref{fig:training_curves}(a) shows the full long-horizon trajectory for the~\MLP~trained on the MNIST subset. The run exhibits the familiar pre-grokking $\to$ grokking transition: training accuracy saturates rapidly, test accuracy improves much later, and the model reaches a high-generalization regime around $10^6$ steps. Continued optimization reveals a third phase. After grokking, test accuracy drops substantially again while training accuracy remains essentially perfect. 
Likewise, as shown in Fig.~\ref{fig:training_losses}, the test loss reaches a minimum at some point. 
This is the regime (green) we call anti-grokking.

Trap count separates this regime from the earlier phases. During pre-grokking, when the model has already fit the training set but has not yet generalized, the average number of detected traps is effectively zero. During grokking, trap count remains near zero. It is only when late-stage anti-grokking begins that trap count rises sharply. This is exactly what one would expect if the traps are measuring departure from a self-averaging random baseline rather than mere scale growth. 

Table~\ref{tab:traps} makes the phase separation explicit. In both the no-weight-decay setting (\texttt{WD=0}) and the weight-decay control (\texttt{WD>0}), traps are absent during pre-grokking and grokking and appear only in anti-grokking. The no-weight-decay run exhibits both a larger trap count and a more severe drop in test accuracy, suggesting that trap count reflects not only onset but also severity.

\subsection{GPT2: the same pattern in a larger architecture}
We also study a GPT2 model designed specifically to induce grokking~\citep{wang2024grokked}, now trained for a very long-horizon to also induce anti-grokking.
The same three-phase structure appears in this very different benchmark. Figure~\ref{fig:training_curves}(c) shows training and test accuracy for the GPT2 transformer together with trap count, this time for each layer. As in the~\MLP, the model first memorizes, then generalizes, and later experiences a substantial degradation in test-accuracy under continued optimization while (mostly) preserving perfect training accuracy.  The appearance of Correlation Traps is also very well correlated with the minimum of the test loss, as shown in Fig.~\ref{fig:training_losses}

\subsection{Weight decay suppresses anti-grokking (trap growth \& overfitting)}
A useful control is the~\MLP~run with nonzero weight decay. Weight decay does not eliminate the three phases entirely, but it reduces both the number of traps and the extent of the late-stage test degradation (Appendix~\ref{app:wd_experiment}; Table~\ref{tab:traps}). This matters for two reasons. First, it shows that anti-grokking is not merely an artifact of the phase definitions: when optimization is regularized, the structural signature is weaker and so is the degradation. Second, it supports the non-self-averaging interpretation; it is not just that the weights are smaller, but that the Correlation Traps are suppressed.

\subsection{Additional validation: perturbation study}
Appendix~\ref{app:l2-perturbation} runs a perturbation experiment with the goal of ruling out a simpler explanation: that trap counts are only tracking the overall size of the weights. We perturb the checkpoints in a way that changes the global weight scale, so norm-based quantities move across training phases. Even under this perturbation, the shuffled-spectrum trap count remains near zero in pre-grokking and grokking, and becomes nonzero only in anti-grokking. Thus, Correlation Traps are not merely a proxy for increasing global weight norm; they are phase-specific and indicative of overfitting.

\section{Trap Classification by a JSD Diagnostic Ablation Test}
\label{sec:kl}

Random Matrix Theory (RMT) lets us detect Correlation Traps, but we also want to know how harmful traps really are; to do this, we define a data-free JSD diagnostic ablation test which can quantify the importance of a trap. Let \(M_{\theta}\) denote the original model and \(M_{\theta\setminus k}\) the model after replacing trap \(k\) with a suitable random vector. Define the temperature-scaled output distributions
\[
p_\theta(\mathbf{x};T)=\operatorname{softmax}\!\left(z_\theta(\mathbf{x})/T\right),
\qquad
p_{\theta\setminus k}(\mathbf{x};T)
=
\operatorname{softmax}\!\left(z_{\theta\setminus k}(\mathbf{x})/T\right).
\]
Given probe inputs \(\mathbf{x}_p\), define the trap-removal score
\begin{equation}
J_k(T)
=
\mathbb{E}_{\mathbf{x}_p}
\!\left[
D_{\mathrm{JS}}
\!\left(
p_\theta(\mathbf{x}_p;T)
\,\middle\|\,
p_{\theta\setminus k}(\mathbf{x}_p;T)
\right)
\right].
\label{eq:js}
\end{equation}
Here \(z_\theta(\mathbf{x})\) and \(z_{\theta\setminus k}(\mathbf{x})\) are the original and trap-removed logits, and \(D_{\mathrm{JS}}\) denotes the Jensen--Shannon divergence
(a symmetric, smoothed version of the KL divergence).

For modular addition and GPT-style models, the probe distribution is random-token sequences. For the MLP, random tokens are not meaningful, so probes are Gaussian inputs matched to a preset mean and standard deviation. $J_k(T)$ measures how important a trap is: large $J_k(T)$ means removing the trap significantly changes the model outputs even on random input data $\mathbf{x}_p$.

To identify a trap as harmful or benign in our experiments, we replace it with a suitable random vector and compute the change ($\Delta$) in test error.  The classification is:
\begin{itemize}[leftmargin=1.2em, itemsep=0pt, topsep=0pt, parsep=0pt]
    \item \textbf{Harmful trap:} replacement changes logits and improves or hurts the test accuracy.
    \item \textbf{Benign trap:} replacement has a negligible effect on the test accuracy.
\end{itemize}
Ablating most traps either causes the $\Delta$ (change in) test accuracy to fall even more when removed / replaced, as well as degrade the training accuracy, or has a very small, sometimes positive effect on one or both.

\begin{figure}[h]
\centering
\begin{minipage}{0.33\columnwidth}
    \centering
  \includegraphics[width=\linewidth]{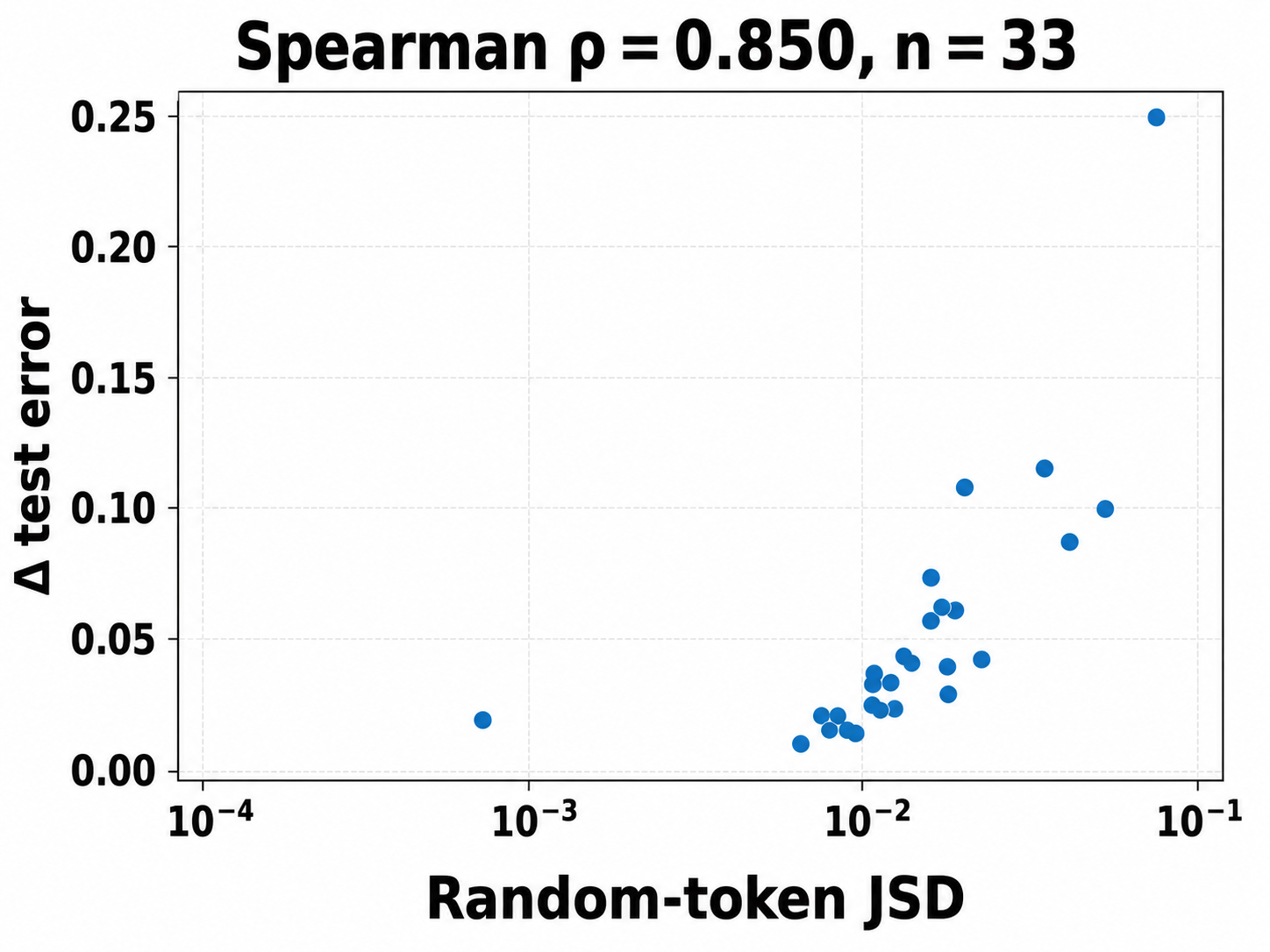}
    \\
    \small (a)~\MLP
\end{minipage}
\begin{minipage}{0.32\columnwidth}
    \centering
    \includegraphics[width=\linewidth]{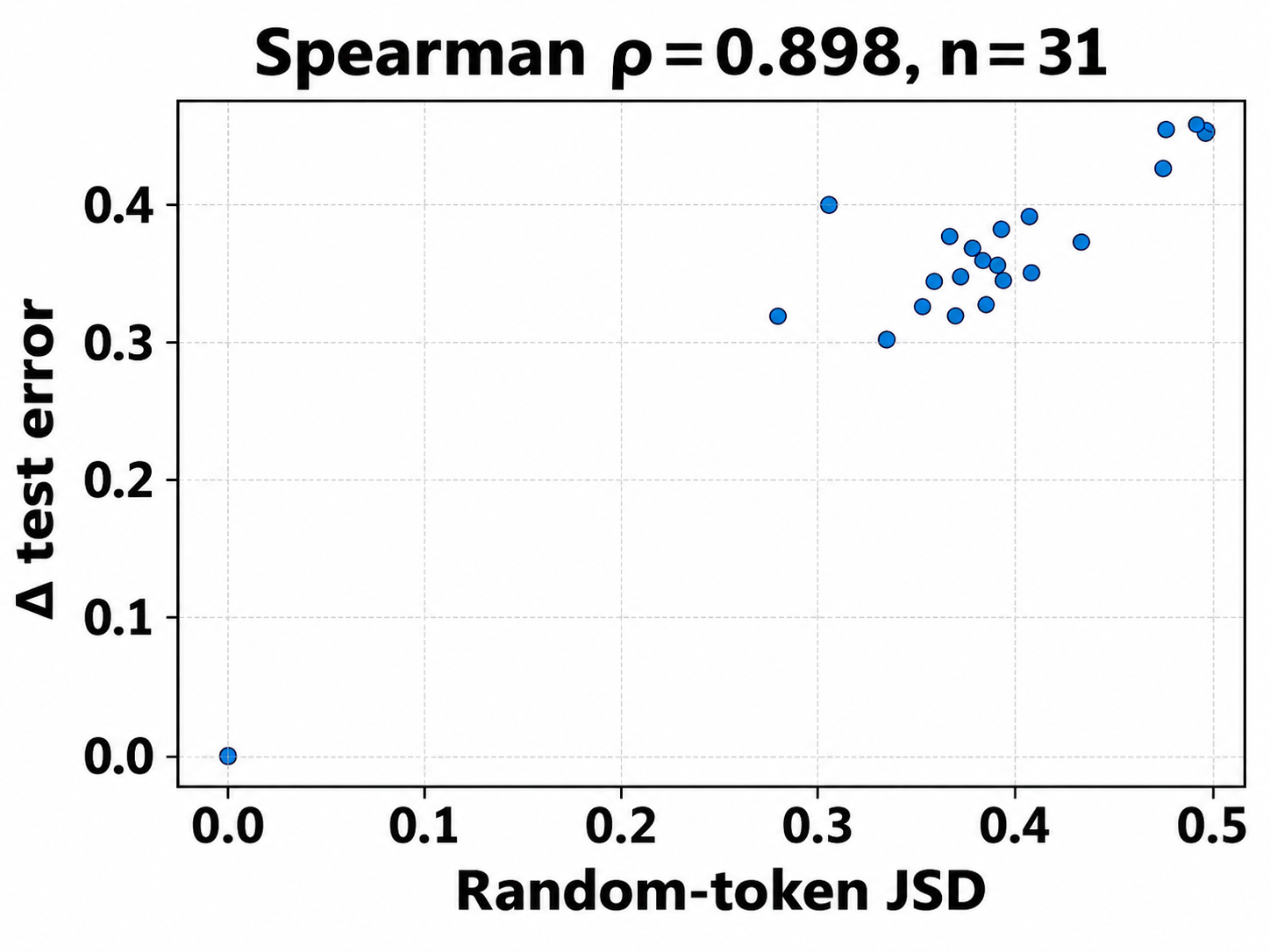}
    \\
    \small (b)~\MA
\end{minipage}
\begin{minipage}{0.32\columnwidth}
    \centering
    \includegraphics[width=\linewidth]{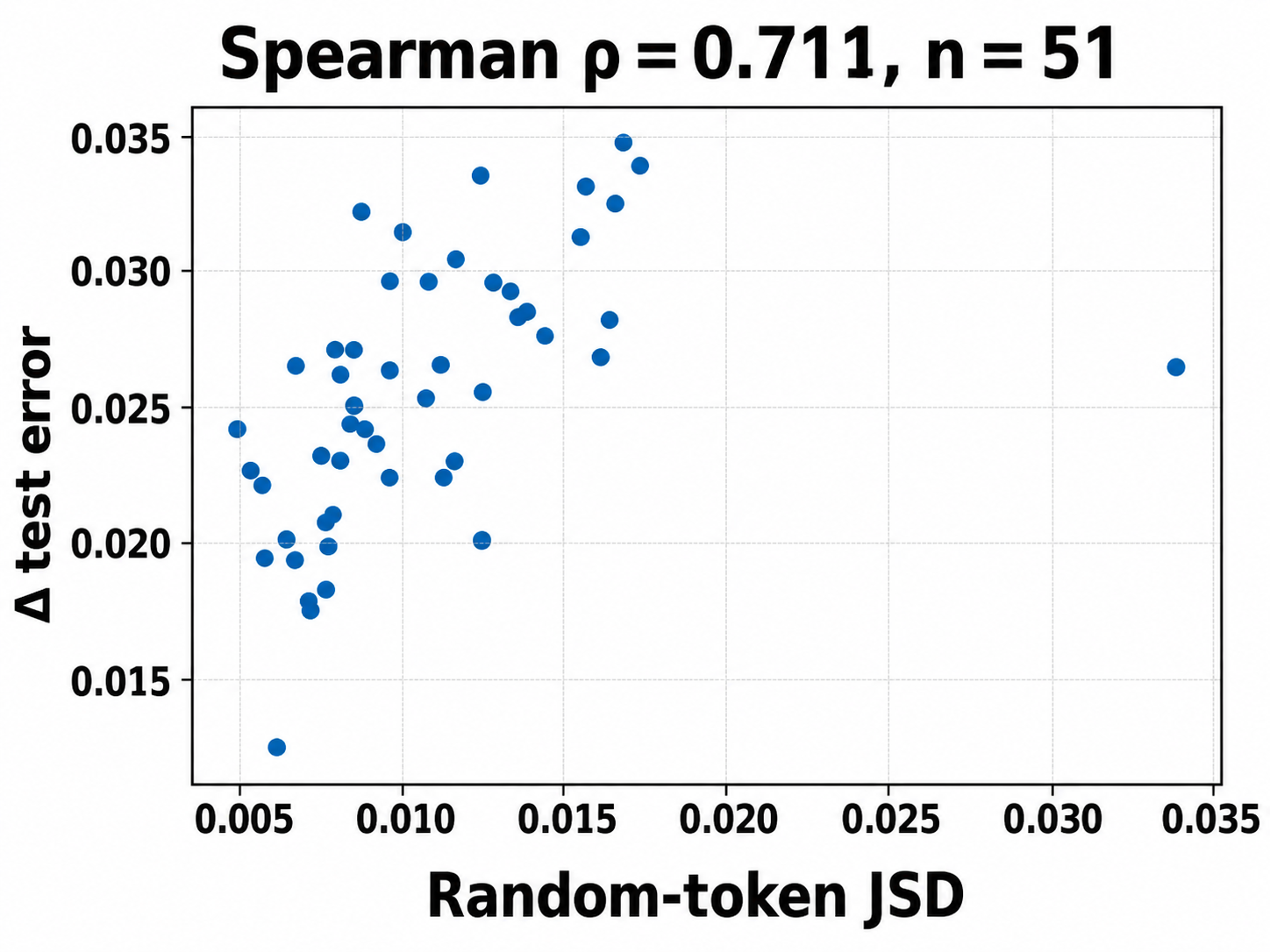}
    \\
    \small (c)~\GPT
\end{minipage}
\caption{\textbf{JSD diagnostic ablation test for selected \MLP, \MA, and \GPT~anti-grokking checkpoints.} Trap removal score $J_k(T=1)$ vs. delta test error for individual traps. Benign traps have $J_k(T=1)\approx 0$ while harmful traps have relatively larger $J_k(T=1)$ scores. 
}
\label{fig:kl-panels}
\end{figure}

Figure~\ref{fig:kl-panels} shows the JSD behavioral diagnostic for the three experiments.
For \MLP~and \MA, we evaluate the first, middle, and final anti-grokking checkpoints.
For \GPT, we analyze a representative  layer from the first anti-grokking checkpoint
($1.5\times10^6$ steps).
Within a given trained model, the trap-removal score $J_k(T=1)$ closely tracks the downstream performance change induced by trap ablation.
The absolute scale depends on the model and trap strength, so JSD is best interpreted as a within-model measure of trap activity,  distinguishing between relatively benign vs. harmful traps.

\section{Broader implications for frontier-scale models}
\label{app:frontier-models}
\begin{figure}[h]
    \centering
    \safeincludegraphics[width=\linewidth]{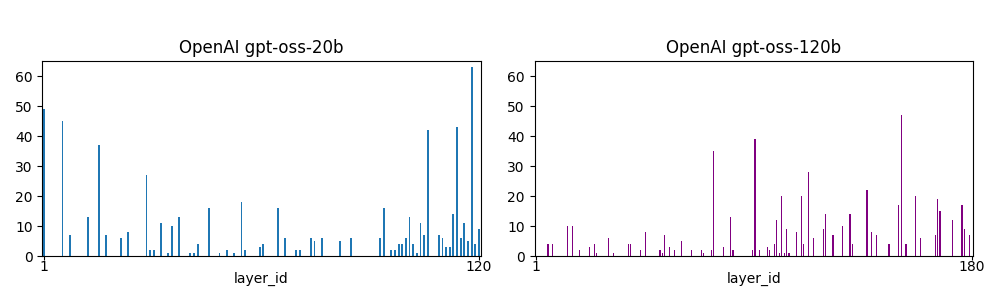}
   \caption{\textbf{Layer-wise Correlation Trap profiles in frontier-scale open-weight LLMs.} Number of detected Correlation Traps per layer for the OpenAI \texttt{gpt-oss-20b} and \texttt{gpt-oss-120b} models.}
    \label{fig:oss_num_traps}
\end{figure}

To motivate the broader relevance of Correlation Traps, we applied the same shuffled-spectrum diagnostic to two open-weight reasoning models released by OpenAI, \texttt{gpt-oss-20b} and \texttt{gpt-oss-120b} \cite{openai_gpt_oss_2025,openai_gpt_oss_model_card_2025}. Figure~\ref{fig:oss_num_traps} reports the number of detected Correlation Traps per layer.

Taken together, these observations suggest that Correlation Traps may provide a practical spectral diagnostic for identifying potentially harmful overfitting structures in frontier-scale foundation models. More broadly, the large number of traps in Figure~\ref{fig:oss_num_traps} suggests that foundation-scale LLMs may be inadvertently overfitting their training data in potentially harmful ways.

\section{Conclusion}

The main lesson is that grokking is not always the end of training. A model can first learn a rule that generalizes, and then lose that generalization under continued optimization. We call this late-stage failure mode \emph{anti-grokking}: specifically, post-generalization, test accuracy degrades, while training accuracy remains high. Across  MNIST~\MLP, Modular Addition (\MA), and GPT2-style grokking experiments, this transition is marked by the appearance of outliers $(\lambda_{trap})$ in the eigenspectrum of one or more randomized, layer-wise weight matrices. We call these outliers \emph{Correlation Traps}.

The term Correlation Trap is operational.  We call these outliers Correlation Traps because they identify latent structures that appear to "trap" the correlations, causing the model to overfit its data,  reducing the test accuracy of the  model, but not (necessarily) the training accuracy.  And as the overfitting gets worse, more traps appear.  Moreover, removing them can even hurt the model's test (and training) accuracy, apparently removing the  correlations the model had learned.

Correlation Traps reveal a structural difference between two superficially similar regimes. In both pre-grokking and anti-grokking, training accuracy can be high while test accuracy is poor. But the weight spectra are different. During pre-grokking, trap counts remain near zero. During successful grokking, they also remain absent or nearly absent. They appear only during anti-grokking; thus, traps do not merely detect memorization, low test accuracy, or high training accuracy. They detect overfitting that arises during anti-grokking.

The diagnostic is simple and weight-only. It does not require training data, test data, gradients, optimizer state, or access to the training pipeline. It only requires a checkpoint. For each layer, we shuffle the weight matrix entry-wise, form the randomized covariance spectrum, fit a Marchenko--Pastur (MP) bulk, and count right-edge outliers beyond the fitted MP/TW edge. The MP/TW law provides a self-averaging baseline. A Correlation Trap is a separated spectral mode that violates this baseline, and may be harmful or benign.

The self-averaging interpretation explains why these modes matter. A trap can disrupt concentration in two ways: geometric localization and/or spectral concentration.  In either case, trap-aligned observables like the test error can become non-self-averaging when the trap overlaps significantly with the  model's generalizing layer eigen-components. This gives at least two concrete mechanisms by which a spectral anomaly in the weights can become a behavioral failure mode.  

We can distinguish harmful from benign traps using a data-free empirical check, called the JSD trap ablation test.  We replace the trap with a random vector, pass probe inputs through the original and ablated models, and measure the Jensen-Shannon divergence between their output distributions. A large JSD means removing the trap significantly changes the model outputs even on random input data, indicative of the effect on real inputs.
The trap is harmful if replacement changes the test accuracy, and benign if it has little-to-no effect.

The \MLP~case study shows what a harmful trap can look like mechanistically. In anti-grokking, mapped first-layer trap directions become prototype-like in pixel space, and intervening on the trap changes the confusion structure. The model still classifies the finite training set correctly, but on trap-affected test examples its behavior increasingly depends on flattened-vector amplitude and prototype statistics, such as \(\Vert x\Vert_2\), rather than the correlated two-dimensional digit pattern. This gives one concrete example of how a model can remain perfectly fit to its training data while losing the correlation-based rule that supported generalization.


The same trap-onset signature is not specific to image MLPs. Modular Addition provides an algorithmic transformer setting where trap counts remain near zero before and during grokking and rise during anti-grokking. The GPT2-style experiment shows the same long-horizon pattern in a transformer language-model architecture. Together, these experiments show that Correlation Traps recur across the model families and task types studied here: dense image classifiers, algorithmic transformers, and GPT-style sequence models.

The frontier-scale screening experiment points to a practical use case. Some large open-weight LLM checkpoints exhibit a large number of traps. These profiles do not by themselves prove harmful overfitting in those models, but they are suggestive of potential problems. For foundation-model training, trap counts could serve as a checkpoint-level warning signal. For fine-tuning, they may help identify base-model layers whose behavior should be tested before adaptation.

We do not claim that every form of overfitting must produce Correlation Traps. The claim here is sharper: in long-horizon grokking, trap onset tracks anti-grokking while pre-grokking provides the trap-free negative control.  Using the JSD diagnostic trap ablation test, traps can be identified as harmful or benign for a given model. These results suggest some forms of harmful overfitting leave detectable signatures in the layer weight matrices of seemingly well-trained models. The next step is to use this diagnostic during training and fine-tuning, and to test whether harmful Correlation Traps can be suppressed, removed, or regularized without damaging useful learned structure.

\section{Limitations}

Our controlled evidence comes from three long-horizon grokking settings: an MNIST~\MLP, modular addition (\MA), and a \GPT-style training trajectory. These experiments expose the pre-grokking, grokking, and anti-grokking phases clearly, but they do not exhaust the space of architectures, optimizers, datasets, losses, or seeds. The GPT-OSS results show that Correlation Traps also occur in large open-weight models, but those results are screening evidence only; they do not by themselves establish harmful behavior. The main risk is over-interpretation: trap counts should not be used as universal model-quality or safety scores without additional probe-based and task-based evaluations.

\paragraph{Potential societal impacts.}
There are over a million open-source model weights available to download, but many of them, however, have no doubt overfit their training data. This study lets users select for models that have fewer signatures of overfitting and potentially better suited to the desired tasks.
Moreover, with the calls for  regulation of foundation-scale models, Correlation Traps can provide regulatory bodies one unbiased way to  gauge of  model safety, and without needing access to proprietary, secure, or even any available data.

\bibliographystyle{plainnat}
\bibliography{references}

\clearpage
\appendix

\section{Experimental Setup and Additional Notes: MLP, MA, and GPT2}
\label{app:exp_setup}

\subsection{MLP Experimental Setup}
We train a Multi-Layer Perceptron (MLP) on a subset of the MNIST dataset using the hyperparameters in Table~\ref{tab:hyperparameters_appendix}. The training subset is constructed by randomly selecting 100 samples from each of the 10 MNIST classes, yielding a balanced dataset of 1{,}000 unique training points. The primary long-horizon runs were performed on a single NVIDIA Quadro P2000 GPU and the main $10^7$-step~\MLP~experiment took approximately 11 hours; a considerable fraction of the wall-clock time is due to checkpointing and saving measurements for later analysis.

\begin{table}[h]
\centering
\caption{MLP experimental hyperparameters used in the study.}
\label{tab:hyperparameters_appendix}
\begin{tabular}{@{}ll@{}}
\toprule
\textbf{Parameter} & \textbf{Value} \\
\midrule
Network Architecture & Fully Connected~\MLP~\\
Depth & 3 Linear layers (Input $\to$ Hidden1 $\to$ Hidden2 $\to$ Output) \\
Width & 200 hidden units per hidden layer \\
Activation Function & ReLU \\
Input Layer Size & 784 (flattened MNIST image $28\times 28$) \\
Output Layer Size & 10 \\
Weight Initialization & Default PyTorch (Kaiming Uniform for weights), parameters scaled by 8.0 \\
Bias Initialization & Default PyTorch (Uniform), then scaled by 8.0 \\
Dataset & MNIST \\
Training Points & 1{,}000 (100 per class, stratified random sampling) \\
Test Points & Standard MNIST test set (10{,}000 samples) \\
Batch Size & 200 \\
Loss Function & MSE with one-hot encoded targets \\
Optimizer & AdamW \\
Learning Rate (LR) & $5\times 10^{-4}$ \\
Weight Decay (WD) & 0.0 (main results), 0.01 (control experiment) \\
AdamW $\beta_1$ & 0.9 \\
AdamW $\beta_2$ & 0.999 \\
AdamW $\epsilon$ & $10^{-8}$ \\
Optimization Steps & $10^7$ \\
Data Type (PyTorch) & \texttt{torch.float64} \\
Random Seed & 0 \\
Software Framework & PyTorch \\
WeightWatcher Version & v0.7.5.5 \cite{weightwatcher} \\
\bottomrule
\end{tabular}
\end{table}

\paragraph{Note on weight decay.}
The primary results in the main paper were obtained with weight decay explicitly set to 0. This isolates the long-horizon optimization dynamics from explicit norm regularization. Runs with nonzero weight decay (\texttt{WD=0.01}) were performed for comparison and exhibit substantially weaker trap growth and milder loss of generalization.

\subsection{MLP Control Experiment with Weight Decay}
\label{app:wd_experiment}

\begin{table}[h]
\centering
\caption{\textbf{Average number of Correlation Traps in~\MLP~with and without Weight Decay (WD).} Trap counts for FC1 and FC2 at representative checkpoints at the right edge of each phase: pre-grokking ($\sim 10^5$ steps), grokking ($10^6$), and anti-grokking ($10^7$). Adding weight decay suppresses trap growth and late-stage collapse, but does not eliminate overfitting entirely.}
\label{tab:traps}
\small
\setlength{\tabcolsep}{5pt}
\begin{tabular}{lccc}
\toprule
\textbf{Setting, Layer} & \textbf{Pre-Grokking} & \textbf{Grokking} & \textbf{Anti-Grokking} \\
\midrule
\texttt{WD=0}, FC1  & $0\pm0$ & $0\pm0$ & $7.5\pm5.6$ \\
\texttt{WD=0}, FC2  & $0\pm0$ & $0\pm0$ & $1\pm0$ \\
\midrule
\texttt{WD>0}, FC1  & $0$ & $0$ & $2.0\pm0.0$ \\
\texttt{WD>0}, FC2  & $0$ & $0$ & $1.0\pm0.0$ \\
\bottomrule
\end{tabular}
\end{table}

The weight-decay control uses the same architecture, data, and optimizer as the main~\MLP~experiment, but with \texttt{WD=0.01}. The central observation is that weight decay suppresses trap growth. As Table~\ref{tab:traps} shows, anti-grokking under \texttt{WD>0} still exhibits traps, but fewer than in the \texttt{WD=0} case, and the associated degradation in test accuracy is much smaller. This supports the interpretation that weight decay mitigates the structural anomalies detected by the shuffled-spectrum diagnostic.

\subsection{Modular Addition (MA) Experiment}
\label{app:modadd}
We additionally evaluate the shuffled-spectrum diagnostic on a modular-addition task with a small one-layer transformer.

\paragraph{Model architecture.}
\begin{table}[h]
\centering
\small
\begin{tabular}{@{}ll@{}}
\toprule
\textbf{Hyper-parameter} & \textbf{Value} \\
\midrule
Layers & 1 \\
Model dimension $d_{\mathrm{model}}$ & 128 \\
MLP hidden size $d_{\mathrm{mlp}}$ & 512 \\
Heads $\times$ head dim & $4\times 32$ \\
Context length $n_{\mathrm{ctx}}$ & 3 \\
Activation & ReLU \\
LayerNorm disabled & \texttt{use\_ln=False} \\
Vocabulary size & 114 ($=$ \texttt{equals\_token}+1) \\
\bottomrule
\end{tabular}
\end{table}

\begin{table}[h]
\centering
\caption{\textbf{Modular Addition phase summary.} Train and test accuracy (mean$\pm$std) across sampled checkpoints in each phase.}
\label{tab:phase_summary_qn}
\begin{tabular}{@{}lcc@{}}
\toprule
\textbf{Phase} & \textbf{Train Acc. (mean$\pm$std)} & \textbf{Test Acc. (mean$\pm$std)} \\
\midrule
Pre-grok & $1.0\pm0.0$ & $0.40\pm0.28$ \\
Grok & $1.0\pm0.0$ & $0.97\pm0.02$ \\
Anti-grok & $1.0\pm0.0$ & $0.68\pm0.11$ \\
\bottomrule
\end{tabular}
\end{table}

\begin{table}[h]
\centering
\caption{\textbf{Modular Addition trap counts.} Number of shuffled-spectrum outliers detected by \WW in each layer and phase.}
\label{tab:layer_traps_qn}
\begin{tabular}{@{}lccc@{}}
\toprule
\textbf{Layer} & \textbf{Pre-grok traps} & \textbf{Grok traps} & \textbf{Anti-grok traps} \\
\midrule
embed.embed & 0.00 & 0.00 & 7 \\
blocks.0.attn.W\_Q & 0.00 & 0.00 & 3 \\
blocks.0.attn.W\_K & 0.00 & 0.00 & 5 \\
blocks.0.attn.W\_V & 0.00 & 0.00 & 3 \\
blocks.0.attn.out\_proj & 0.00 & 0.00 & 4 \\
blocks.0.mlp.fc1 & 0.00 & 0.00 & 5 \\
blocks.0.mlp.fc2 & 0.00 & 0.00 & 7 \\
unembed.unembed & 0.00 & 0.00 & 5 \\
\midrule
\textbf{Mean$\pm$Std} & $\mathbf{0.00\pm0.00}$ & $\mathbf{0.00\pm0.00}$ & $\mathbf{4.88\pm1.55}$ \\
\bottomrule
\end{tabular}
\end{table}

Trap counts are effectively zero in pre-grokking and grokking and rise sharply across many layers in anti-grokking. This reproduces the same qualitative signature observed in the~\MLP: trap-free shuffled spectra in the first two phases and numerous shuffled-spectrum outliers during late-stage collapse.

\subsection{GPT2 Experimental Setup}
\label{app:gpt2}

We evaluate the shuffled-spectrum diagnostic on the synthetic composition benchmark from
GrokkedTransformer~\citep{wang2024grokked}. The task is generated from a random knowledge
graph of atomic one-hop facts and latent two-hop composition rules. Table~\ref{tab:gpt2_dataset}
summarizes the dataset construction, Table~\ref{tab:gpt2_splits} defines the evaluation splits,
and Table~\ref{tab:gpt2_training} lists the training and analysis configuration.

\begin{table}[t]
\centering
\caption{\textbf{GPT2 synthetic composition dataset.} The benchmark is generated from a random
knowledge graph with atomic facts $(h,r,t)$, queried as $(h,r)\mapsto t$, and inferred two-hop
queries $(h,r_1,r_2)\mapsto t$ induced by pairs of atomic facts $(h,r_1,m)$ and $(m,r_2,t)$.}
\label{tab:gpt2_dataset}
\begin{tabular}{ll}
\toprule
\textbf{Dataset parameter} & \textbf{Value} \\
\midrule
Dataset variant & \texttt{composition.2000.200.12.6} \\
Number of entities & $2000$ \\
Number of relations & $200$ \\
Atomic outgoing facts per entity & $20$ \\
Atomic fact form & $(h,r,t)$ \\
Atomic query form & $(h,r)\mapsto t$ \\
Composition rule & $(h,r_1,m), (m,r_2,t) \Rightarrow (h,r_1,r_2)\mapsto t$ \\
Atomic partition & $\mathrm{atomic}_{\mathrm{ID}} \cup \mathrm{atomic}_{\mathrm{OOD}}$ \\
OOD atomic fraction & Approximately $5\%$ \\
Training inferred downsampling & $12.6 \times |\mathrm{atomic}_{\mathrm{ID}}|$ \\
\bottomrule
\end{tabular}
\end{table}

\begin{table}[t]
\centering
\caption{\textbf{GPT2 train/test split definitions.} The main grokking curve treats
$\mathrm{train\_inferred}_{\mathrm{ID}}$ as the training curve and
$\mathrm{test\_inferred}_{\mathrm{ID}}$ as the test curve. The OOD inferred split is reserved as
a stronger systematicity diagnostic.}
\label{tab:gpt2_splits}
\begin{tabular}{lll}
\toprule
\textbf{Split} & \textbf{Included in training?} & \textbf{Role in evaluation} \\
\midrule
$\mathrm{id\_atomic}$ & Yes & ID atomic memorization/control accuracy \\
$\mathrm{ood\_atomic}$ & Yes & OOD atomic memorization/control accuracy \\
$\mathrm{train\_inferred}_{\mathrm{ID}}$ & Yes & Main ``train'' curve for inferred facts \\
$\mathrm{test\_inferred}_{\mathrm{ID}}$ & No & Main ``test'' curve for ID compositional generalization \\
$\mathrm{test\_inferred}_{\mathrm{OOD}}$ & No & Stronger systematicity diagnostic \\
\bottomrule
\end{tabular}
\end{table}

The training set contains all atomic facts, both ID and OOD, together with a random subset of
inferred facts deduced only from $\mathrm{atomic}_{\mathrm{ID}}$. Therefore, the primary
train/test comparison measures in-distribution compositional generalization: the model must apply
the learned latent composition rule to unseen inferred facts drawn from the same atomic pool used to
generate the training compositions. We report $\mathrm{test\_inferred}_{\mathrm{OOD}}$ separately,
since it probes generalization to inferred facts involving the held-out atomic pool rather than the
main grokking transition.

\begin{table}[t]
\centering
\caption{\textbf{GPT2 model, training, and evaluation configuration.} The model is an
8-layer GPT2-style decoder-only transformer initialized from scratch and trained with the
GrokkedTransformer training code.}
\label{tab:gpt2_training}
\begin{tabular}{ll}
\toprule
\textbf{Hyper-parameter} & \textbf{Value} \\
\midrule
Architecture & GPT2-style decoder-only transformer \\
Initialization & From scratch \\
Number of layers & $8$ \\
Vocabulary & Synthetic entity, relation, and answer-delimiter tokens added \\
Maximum sequence length & $10$ \\
Batch size & $512$ \\
Optimizer & AdamW \\
Learning rate & $10^{-4}$ \\
Weight decay & $0.001$ \\
Schedule & Constant schedule with warmup \\
Precision & Mixed precision \\
Training steps & $1.5\times 10^6$ \\
Checkpointing & Checkpoints saved throughout training \\
Evaluation protocol & Exact-match generation \\
Correctness criterion & Generated answer token matches target before closing delimiter \\
Reported accuracies &
$\mathrm{id\_atomic}$, $\mathrm{ood\_atomic}$,
$\mathrm{train\_inferred}_{\mathrm{ID}}$,
$\mathrm{test\_inferred}_{\mathrm{ID}}$,
$\mathrm{test\_inferred}_{\mathrm{OOD}}$ \\
\bottomrule
\end{tabular}
\end{table}

For spectral and ablation analyses, we run WeightWatcher on the transformer block matrices and
exclude the positional embedding layer \texttt{wpe}. This exclusion avoids treating the positional
embedding table as a standard learned linear map in the layerwise trap analysis.

\begin{table}[t]
\centering
\caption{\textbf{GPT2 spectral-analysis convention.} WeightWatcher is applied to learned
transformer block matrices, with positional embeddings excluded from the layerwise trap count.}
\label{tab:gpt2_spectral}
\begin{tabular}{ll}
\toprule
\textbf{Analysis choice} & \textbf{Convention} \\
\midrule
Diagnostic & Shuffled-spectrum outlier detection with WeightWatcher \\
Included modules & Transformer block matrices \\
Excluded module & Positional embedding layer \texttt{wpe} \\
Reason for exclusion & \texttt{wpe} is an embedding table, not a standard learned linear map \\
Layerwise output & Trap counts per analyzed transformer matrix \\
\bottomrule
\end{tabular}
\end{table}

\section{Mechanistic Case Study on Anti-Grokking in the MLP Experiment}
The randomized trap itself is only a diagnostic of atypical amplitude disorder. To interpret the failure mode, we map the trap mode back to the original unshuffled layer coordinates using the \WW procedure summarized below and visualized in Fig. \ref{fig:v1-pixel}. In other words, the trap tells us that a particular direction has moved into an overfit sector of weight space; the prototype-like structure appears when that direction is expressed back in the coordinates of the original layer.


\subsection{Localized leading directions induce norm-based prototype collapse}
\label{sec:norm_probe}

\begin{figure}[t]
    \centering
    \includegraphics[width=\linewidth]{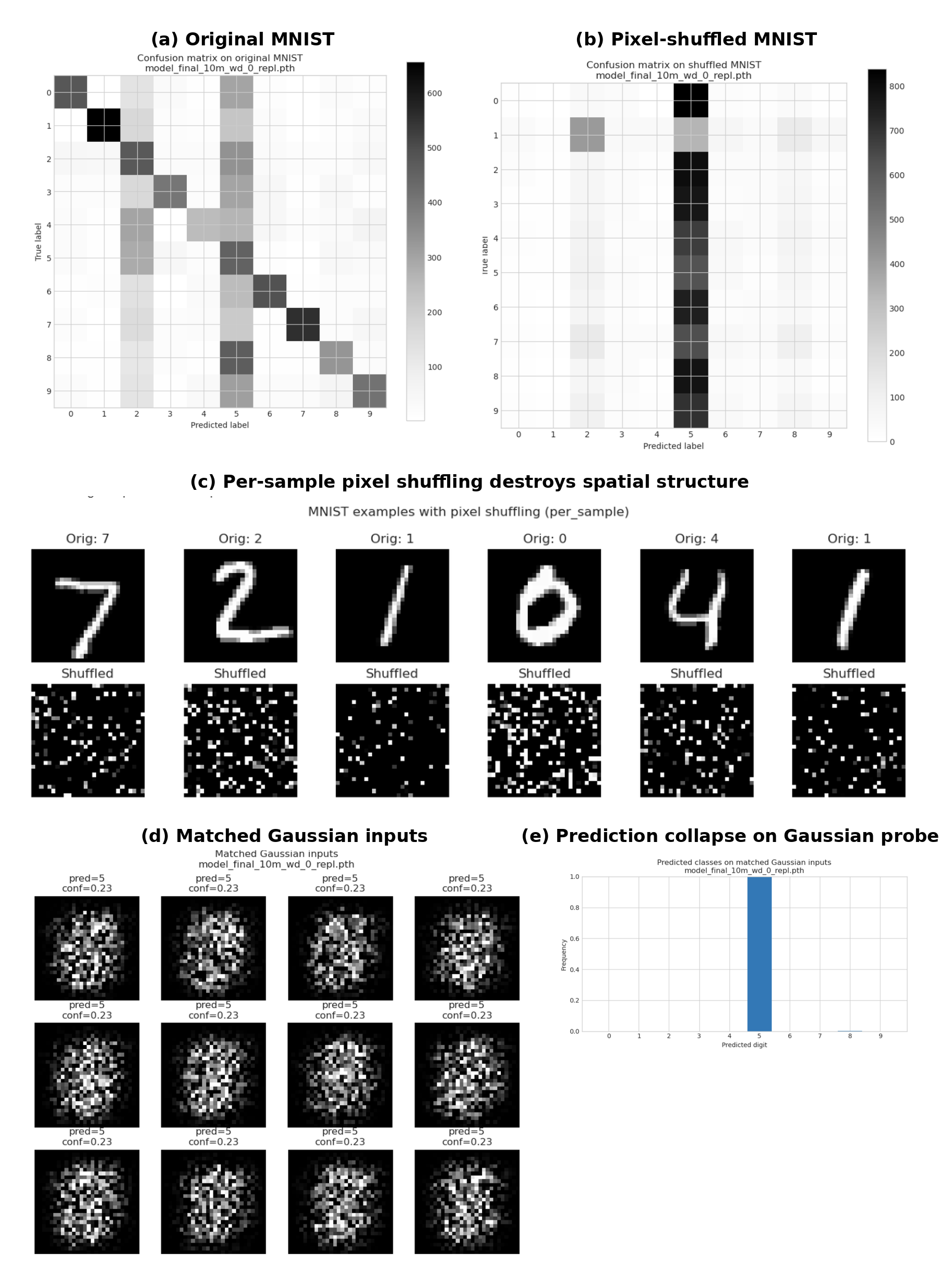}
    \caption{
    Evidence for norm-based prototype collapse in the anti-grokking \MLP{}.
    (a) The original MNIST confusion matrix remains partially structured.
    (b--c) After per-sample pixel shuffling, spatial digit information is destroyed, but predictions collapse toward a small set of labels.
    (d--e) Matched Gaussian probes contain no digit structure, yet the model predicts almost entirely class $5$ with nearly constant confidence.
    Together, these probes indicate that late-stage predictions are governed by global input magnitude rather than stable digit semantics.
    }
    \label{fig:norm-probe}
\end{figure}

Figure~\ref{fig:norm-probe} shows that the anti-grokking \MLP{} no longer uses stable digit geometry in the usual sense. On original MNIST, the confusion matrix remains structured, but when each image is independently pixel-shuffled, the same network collapses most predictions into a small number of classes, especially class $5$. This means the classifier is not simply recognizing spatial digit templates. Instead, it is responding to low-order global statistics that survive shuffling, most notably the overall input scale.

\begin{figure}[h]
  \centering
  \begin{subfigure}[b]{0.32\columnwidth}
    \centering
    \safeincludegraphics[width=\linewidth]{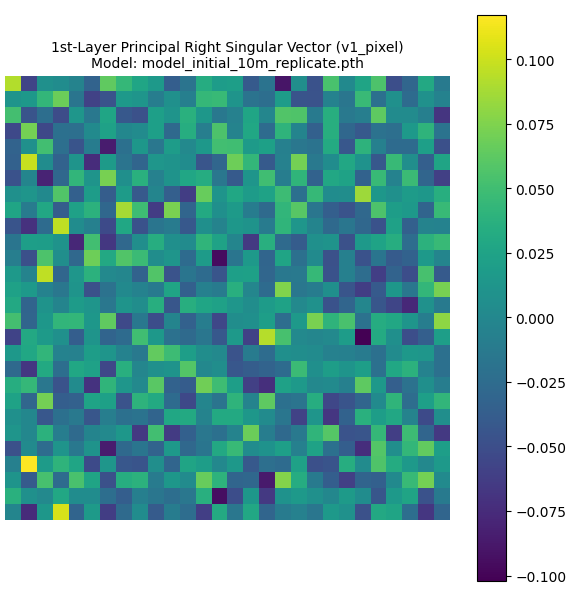}
    \caption{Pre-grokking}
  \end{subfigure}
  \begin{subfigure}[b]{0.32\columnwidth}
    \centering
    \safeincludegraphics[width=\linewidth]{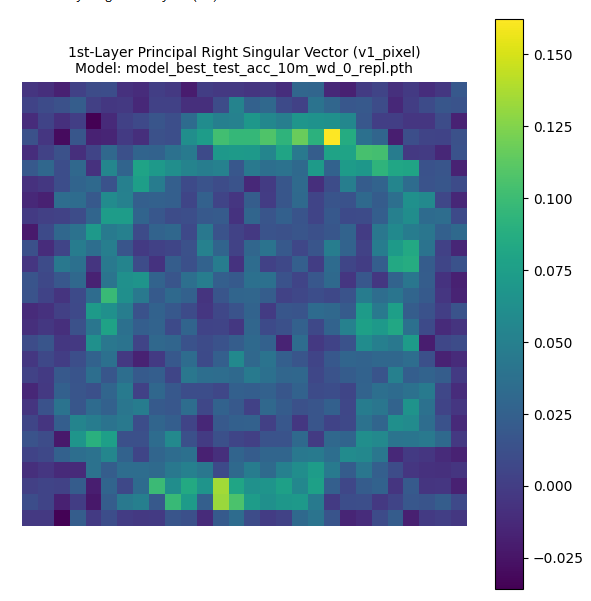}
    \caption{Grokking}
  \end{subfigure}
  \begin{subfigure}[b]{0.32\columnwidth}
    \centering
    \safeincludegraphics[width=\linewidth]{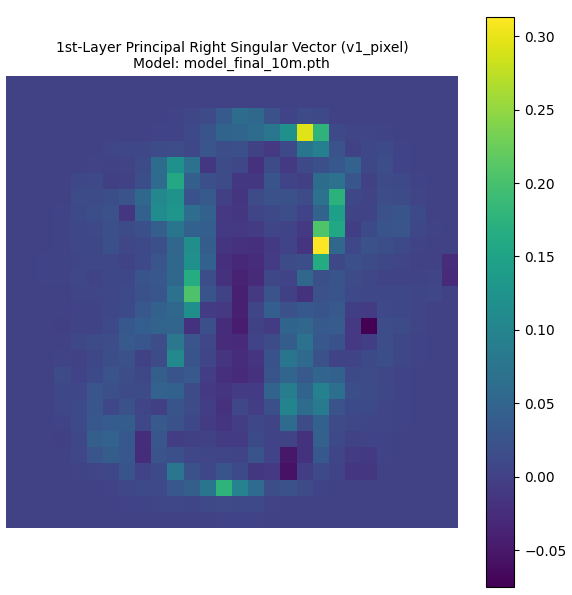}
    \caption{Anti-grokking}
  \end{subfigure}
  \caption{\textbf{Prototype-like directions associated with trapped layers.} Largest right singular vector $\mathbf{v}^{(1)}$ of $\mathbf{W}_1$ in pixel space. The corresponding direction evolves from unstructured noise in pre-grokking, to a smooth global template during grokking, to a localized prototype-like image in anti-grokking. By mapping the trap back to the FC1 layer, one can observe  the overfit sector $\mathbf{W}_1$.}
  \label{fig:v1-pixel}
\end{figure}

To isolate this effect, we next remove digit structure entirely and probe the model with matched Gaussian inputs whose mean and variance match the MNIST input distribution. These inputs contain no class information and no coherent strokes, yet the model again predicts almost exclusively class $5$, with nearly constant confidence. Thus, the late-stage classifier has developed a degenerate decision rule in which prediction is strongly controlled by $\|\mathbf{x}\|_2$-like input magnitude rather than semantic digit shape. This supports the interpretation in Figure~\ref{fig:v1-pixel}: localized leading directions become prototype-like but uncorrelated with the true task. They can preserve perfect training accuracy while inducing brittle, norm-sensitive behavior on perturbed or data-free probes.

This is the sense in which the trap diagnoses harmful overfitting without being the learned pattern itself. The shuffled trap says that the randomized layer has become atypical. When we pull that mode back into the original layer, we see which direction in the true weight matrix has condensed into an overfit prototype, but based solely on the statistical properties of the test instance elements $x^{test}_{i,j}$, such as the test instance norm $\Vert\mathbf{x}^{test}\Vert_2$,  and not their correlated pattern structure.

\subsection{Intervening on the trap direction changes behavior}
\begin{figure}[t]
  \centering
  \safeincludegraphics[width=0.80\textwidth]{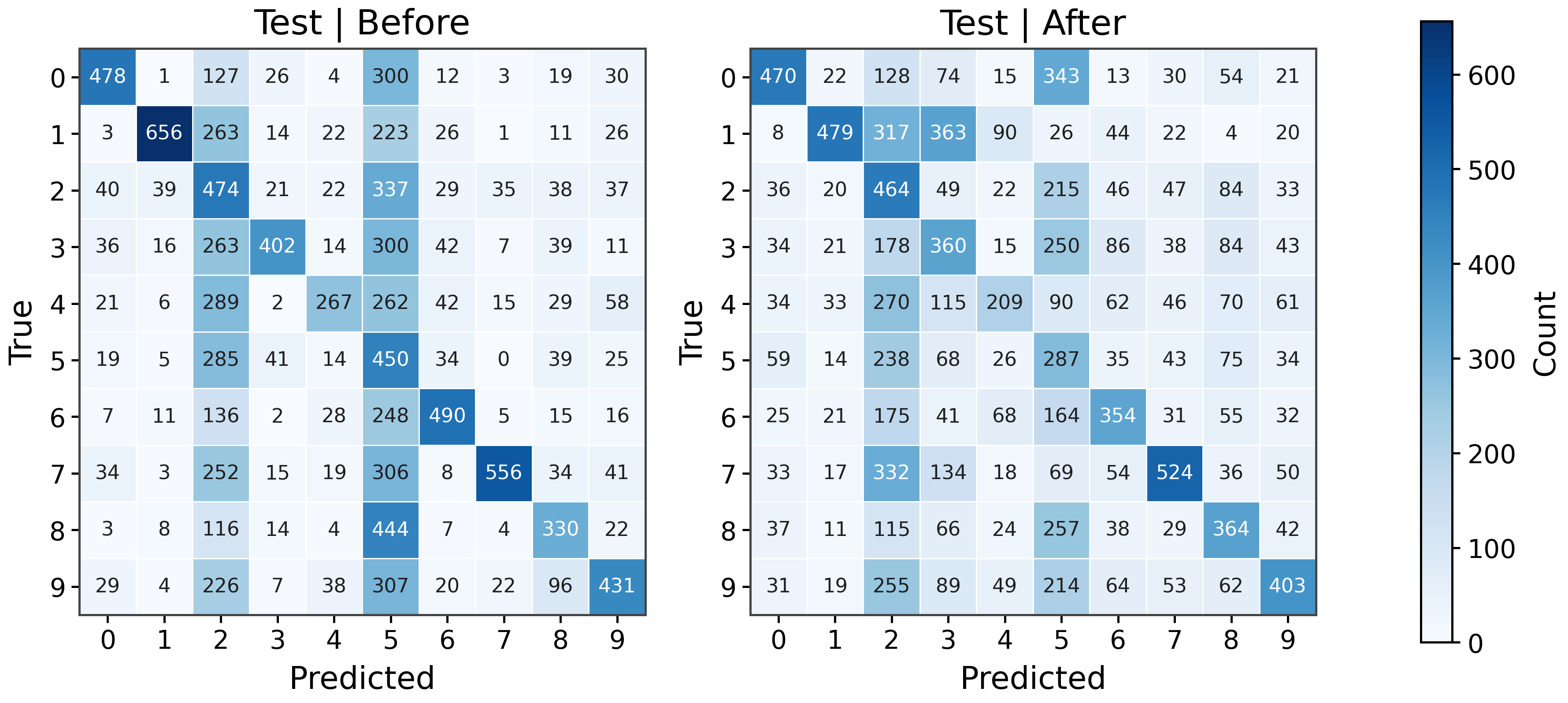}
  \caption{\textbf{Trap intervention in the anti-grokking phase.} (a) Confusion matrix for the full model in anti-grokking. (b) Confusion matrix after replacing the FC1 trap direction with a matched Gaussian random vector. The strong bias toward one prototype class (i.e., $5$) is substantially reduced.}
  \label{fig:confusion_matrices}
\end{figure}

Figure~\ref{fig:confusion_matrices} shows that this structural interpretation is not merely visual. In the anti-grokking regime, an FC1 trap biases many test examples toward a prototype class. Replacing that trap direction with a matched random vector changes the confusion structure and weakens the prototype bias. This is important: it shows that the overfit sector identified through the trap is behaviorally active.

\subsection{Traps recur under continued optimization}
When a localized trap is removed, it typically reappears under continued training. The same data, objective, and optimizer continue to favor that condensed direction. This recurrence is consistent with the idea that the trap is an emergent finite-size instability of the long-horizon optimization dynamics rather than a static artifact of one checkpoint.

\subsection{Leading Eigenvectors, Receptive Fields, and Structural Outliers}
\label{sec:leading_vectors}
For a weight matrix $\mathbf{W}\in\mathbb{R}^{N\times M}$, write
\begin{equation}
\mathbf{X}(\mathbf{W}) := \frac{1}{N}\mathbf{W}^\top \mathbf{W},
\qquad
\lambda_{\max}(\mathbf{W}) := \lambda_{\max}\bigl(\mathbf{X}(\mathbf{W})\bigr).
\end{equation}
If $\mathbf{W}=U\boldsymbol{\Sigma} \mathbf{V}^\top$ is the thin SVD, then the eigenpairs of $\mathbf{X}(\mathbf{W})$ are $(\sigma_k^2/N,\mathbf{v}_k)$, so the leading right singular vector identifies the dominant direction in the input space of the layer. In the~\MLP, this makes the first-layer leading vector directly visualizable in pixel coordinates.

\begin{figure*}[t]
  \centering
  \begin{subfigure}[b]{0.32\textwidth}
    \centering
    \safeincludegraphics[width=\linewidth]{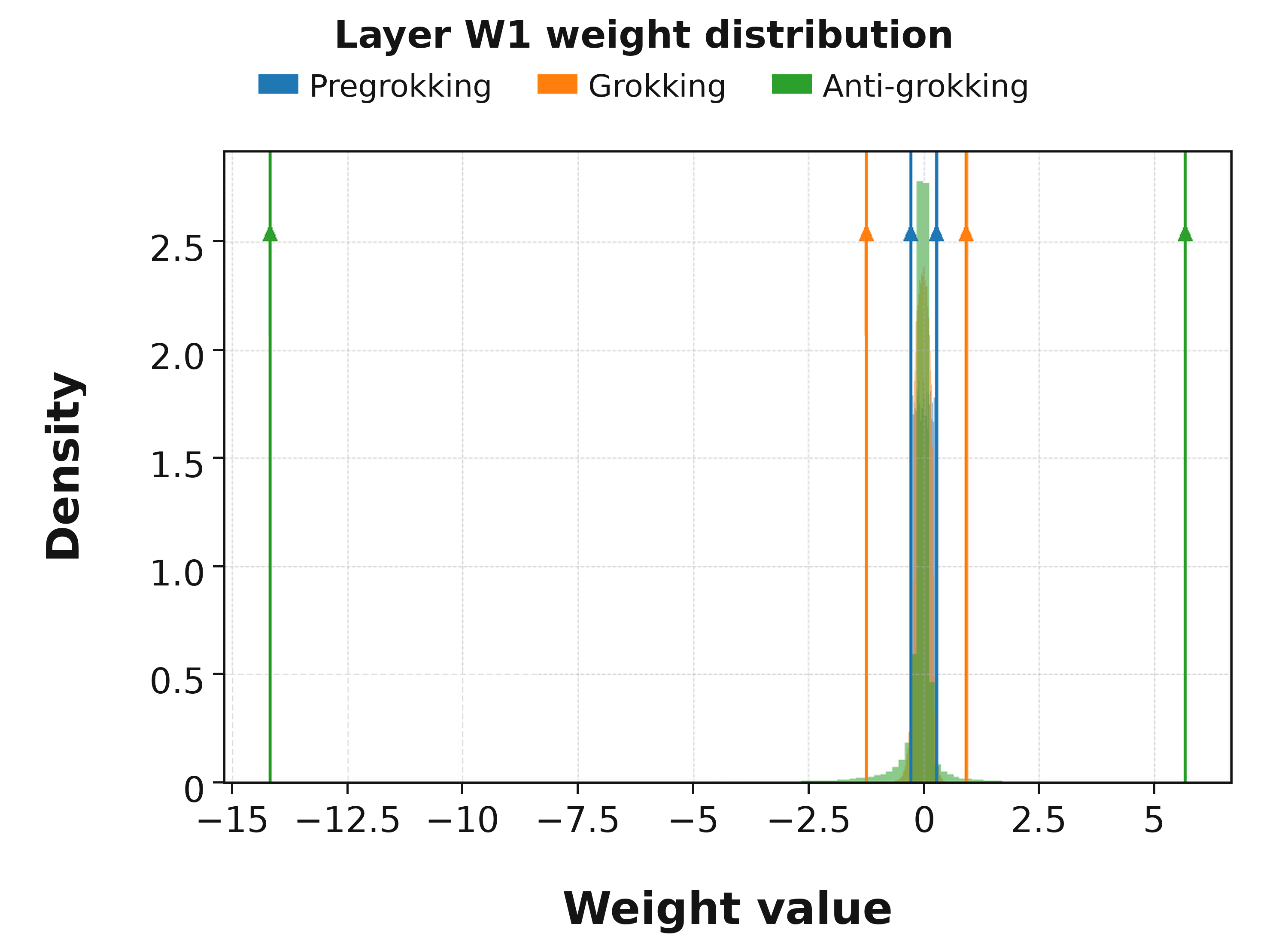}
    \caption{$\mathbf{W}_1$ weights}
    \label{fig:w1-hist}
  \end{subfigure}\hfill
  \begin{subfigure}[b]{0.32\textwidth}
    \centering
    \safeincludegraphics[width=\linewidth]{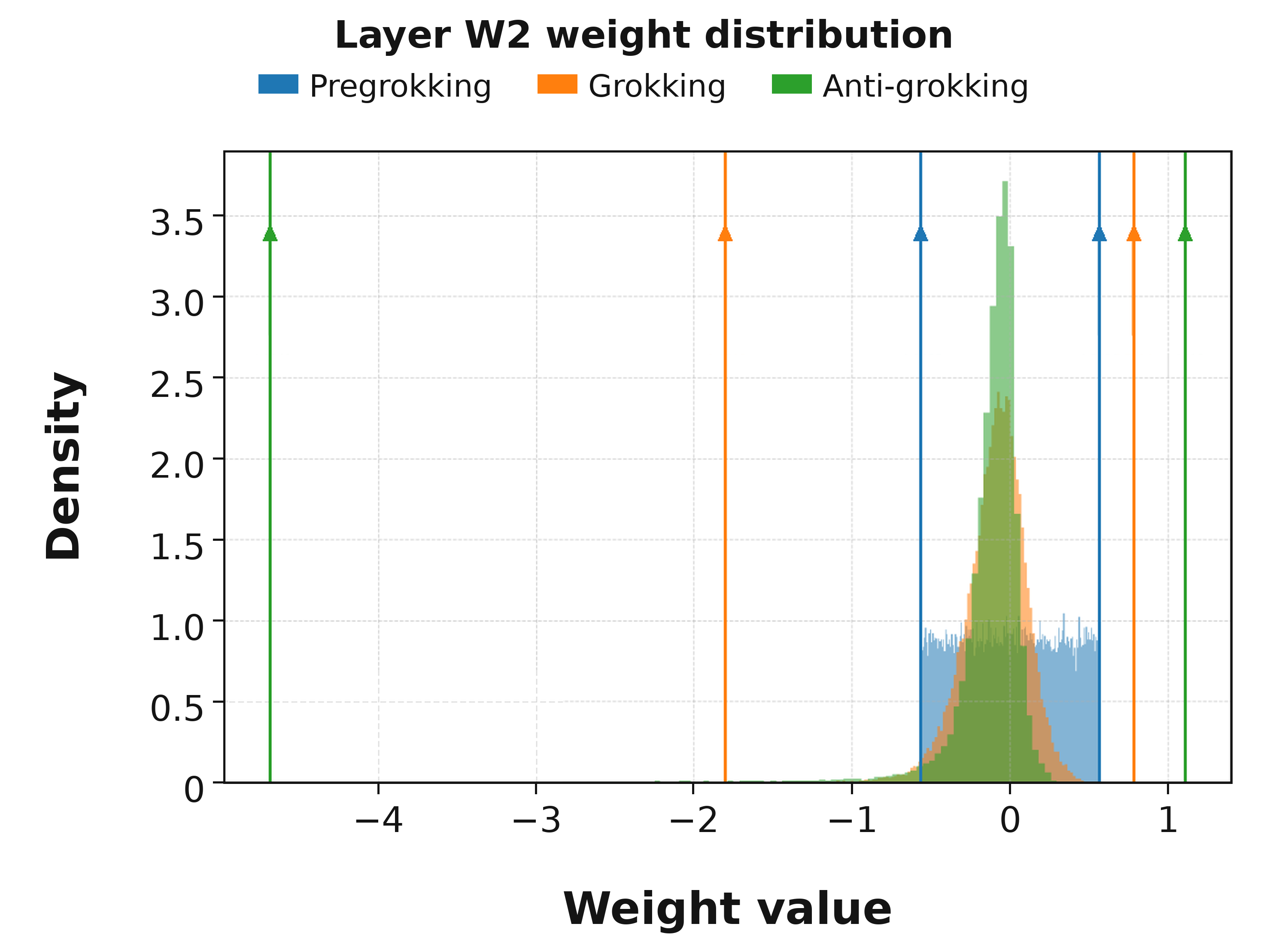}
    \caption{$\mathbf{W}_2$ weights}
    \label{fig:w2-hist}
  \end{subfigure}\hfill
  \begin{subfigure}[b]{0.32\textwidth}
    \centering
    \safeincludegraphics[width=\linewidth]{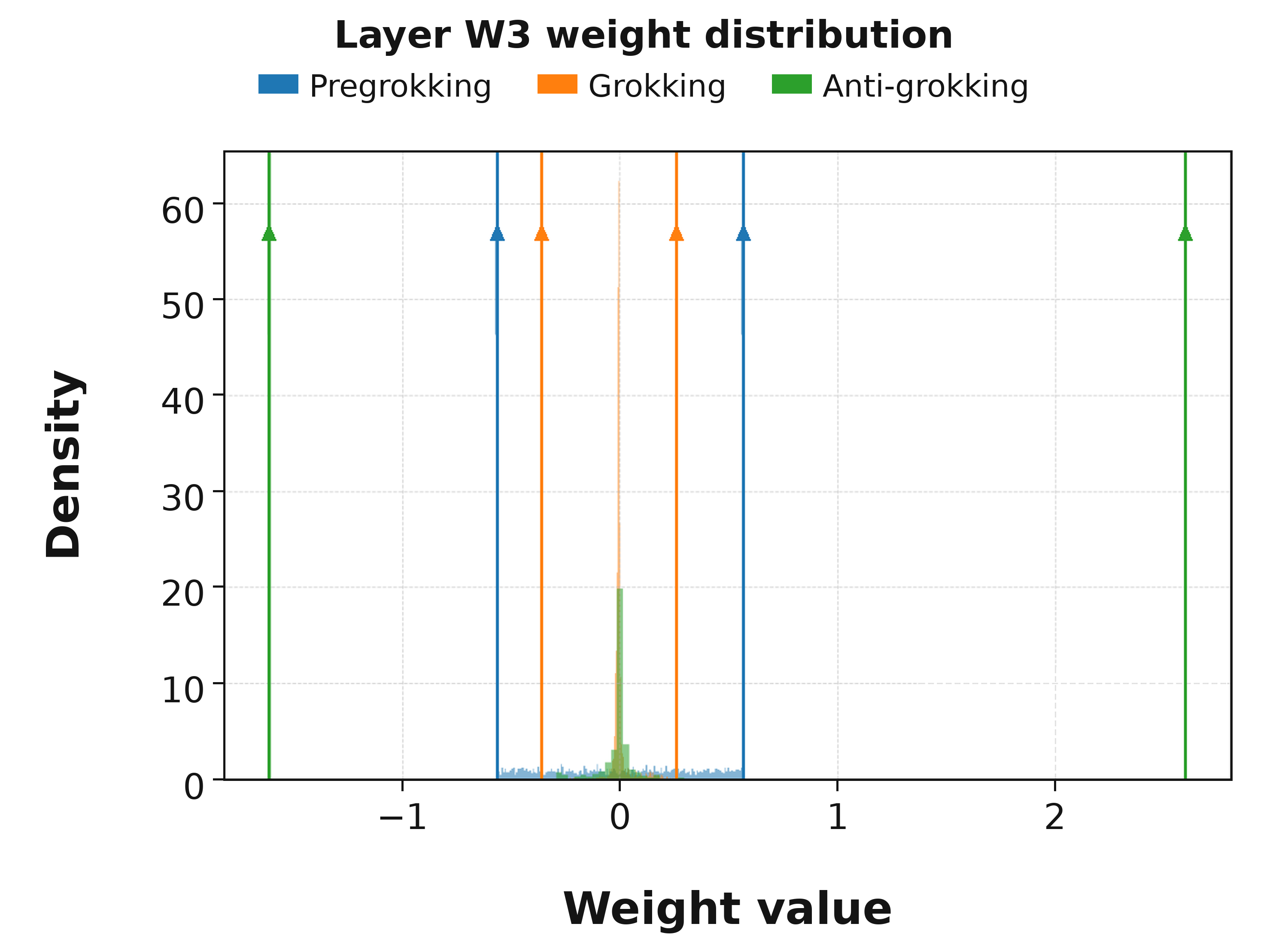}
    \caption{$\mathbf{W}_3$ weights}
    \label{fig:w3-hist}
  \end{subfigure}
  \caption{\textbf{Weight distributions with extreme coordinates across checkpoints.} Anti-grokking introduces structural outliers into the weight values of all three layers. These extreme coordinates are consistent with the shuffled-spectrum outliers counted as Correlation Traps.}
  \label{fig:w-hists}
\end{figure*}

Figure~\ref{fig:w-hists} shows that anti-grokking introduces visibly extreme coordinates into the weight distributions of all three layers. This supports the idea that the late-stage trapped regime is associated with large-magnitude perturbations, not just diffuse reshaping of the bulk.

\begin{figure*}[t]
  \centering
  \begin{subfigure}[b]{1.00\textwidth}
    \centering
    \safeincludegraphics[width=\linewidth]{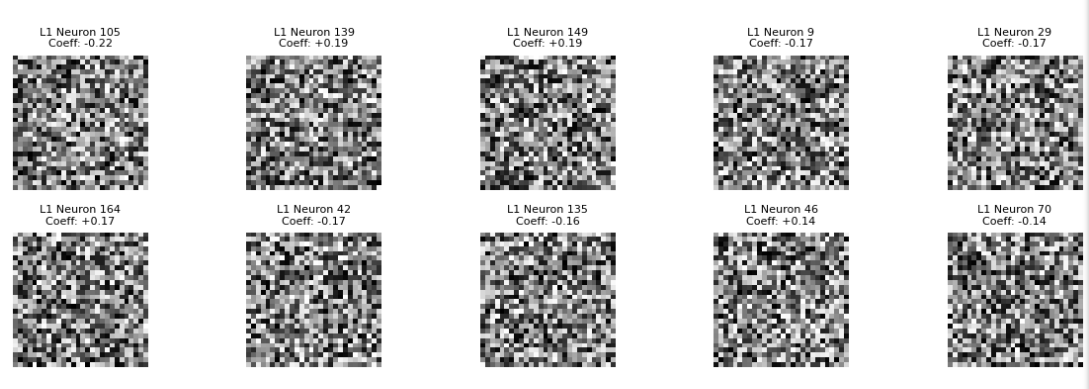}
    \caption{Top-10 rows of $\mathbf{W}_1$ (pre-grokking)}
  \end{subfigure}\hfill
  \begin{subfigure}[b]{1.00\textwidth}
    \centering
    \safeincludegraphics[width=\linewidth]{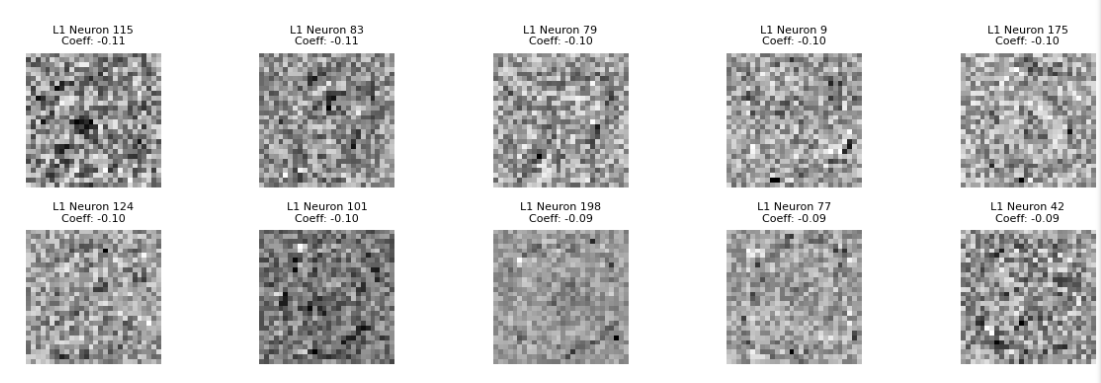}
    \caption{Top-10 rows of $\mathbf{W}_1$ (grokking)}
  \end{subfigure}\hfill
  \begin{subfigure}[b]{1.00\textwidth}
    \centering
    \safeincludegraphics[width=\linewidth]{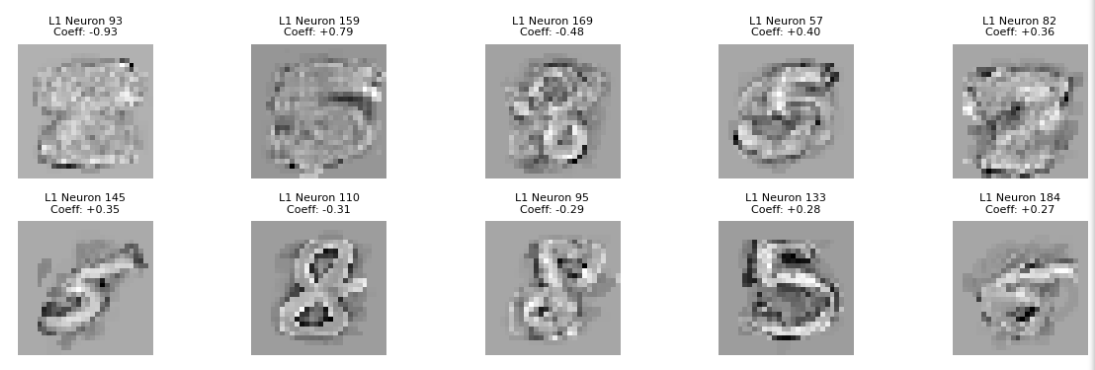}
    \caption{Top-10 rows of $\mathbf{W}_1$ (anti-grokking)}
  \end{subfigure}
  \caption{\textbf{Rows of $\mathbf{W}_1$ selected by leading-vector localization.} The selected receptive fields are noise-like in pre-grokking, remain diffuse around peak generalization, and become recognizable digit templates in anti-grokking.}
  \label{fig:l1-rfs}
\end{figure*}

Figure~\ref{fig:l1-rfs} further sharpens the mechanism. Rows of $\mathbf{W}_1$ selected by leading-vector localization evolve from noise-like patterns to digit-like receptive fields in anti-grokking. This is consistent with the picture developed in the main text: harmful traps are associated with sharply localized prototype directions. In the notation of the paper, these same rows correspond to large $\overlapl{\mathbf{v}_k}$ values when the mapped trap direction is projected onto the dominant eigenvectors of $\mathbf{X}(\mathbf{W}_1)$.

\FloatBarrier

\subsection{Is the distribution of weights in anti-grokking simply very heavy-tailed?}

If overfitting is driven by atypically large entries, one could imagine inspecting the largest weights directly. In practice this is unsatisfactory. A threshold on the largest entries is necessarily ad hoc, depends on scale and architecture, and cannot distinguish diffuse large-weight growth from a genuinely condensed harmful mode. Direct tail fitting is also problematic. The element histogram can look roughly Laplacian in one regime, power-law-like in another, and often shows finite-size crossovers or mixture behavior rather than a clean single exponent. Standard procedures require choosing quantiles or cutoffs, and with millions of squared elements even brute-force fitting can be numerically unstable and conceptually ambiguous.

Instead we use the MP law as a nonparametric self-averaging baseline for the entry-wise randomized layer. The question becomes simple: after the learned arrangement is destroyed, how far does the randomized spectrum still sit outside the MP bulk? This turns the detection problem into one of atypical spectral mass rather than unstable histogram fitting. The MP null is not only a statement about eigenvalues. In the random-matrix regime the eigenvalue density concentrates into a deterministic bulk and the associated eigenvectors are mostly delocalized and look random uniform. A shuffled layer that develops a separated right-edge eigenvalue or a visibly localized top vector has therefore departed from that null. In this sense, good MP fits are empirical evidence of self-averaging for the randomized layer, and strong deviations from the fit are qualitative evidence of non-self-averaging.

The point is not that $\mathbf{W}_{ij}$ must follow a very heavy-tailed law. Empirically the entries can look roughly Laplacian, power-law-like with exponent $\alpha\gtrsim2$, or like a finite-size mixture. Our theory therefore uses replacement-compatible probes of the learned trap vectors rather than a single asymptotic tail model. The mechanism does not require exponent $<2$, and the observed traps are often driven by several large or moderately large entries rather than one extraordinary coordinate.

\FloatBarrier

\section{MLP Perturbation Study} 
\label{app:l2-perturbation}

\paragraph{Do Traps Simply Reflect the Scale of the Weights?} As an additional robustness check, we want to eliminate the possibility that trap counts are merely tracking the scale of the weights.
We evaluate the number of traps on a perturbed version of the MLP checkpoints. The perturbation changes the overall scale of the weights, so global norm-based quantities move across training phases.  Even under this perturbation, trap counts remain near zero in pre-grokking and grokking, and become nonzero only in anti-grokking.

\paragraph{Protocol and phase definitions.}
For this study we use phase boundaries pre-grok ($10^2$--$5\times10^4$ steps), grok ($5\times10^4$--$5\times10^5$ steps), and anti-grok ($>5\times10^5$ steps). The reported statistics are mean$\pm$std over sampled checkpoints within each phase.

\begin{table}[H]
\centering
\caption{Accuracy by phase under perturbation.}
\label{tab:app-acc-by-phase}
\begin{tabular}{lcc}
\toprule
Phase & Train Accuracy (mean$\pm$std) & Test Accuracy (mean$\pm$std) \\
\midrule
Pre-grok & $0.7298 \pm 0.0408$ & $0.4926 \pm 0.0187$ \\
Grok & $1.0000 \pm 0.0000$ & $0.8879 \pm 0.0051$ \\
Anti-grok & $1.0000 \pm 0.0000$ & $0.6184 \pm 0.1086$ \\
\bottomrule
\end{tabular}
\end{table}

\begin{table}[H]
\centering
\caption{Global $\ell_2$ norm by phase under perturbation.}
\label{tab:app-l2-by-phase}
\begin{tabular}{lc}
\toprule
Phase & WeightNorm (mean$\pm$std) \\
\midrule
Pre-grok & $17.52 \pm 0.00$ \\
Grok & $16.80 \pm 0.22$ \\
Anti-grok & $27.64 \pm 8.39$ \\
\bottomrule
\end{tabular}
\end{table}

\begin{table}[H]
\centering
\caption{Layer-wise trap statistic by phase under perturbation.}
\label{tab:app-corr-traps}
\begin{tabular}{lccc}
\toprule
Layer & Pre-grok & Grok & Anti-grok \\
\midrule
1 & $0.00$ & $0.00$ & $7.00$ \\
3 & $0.00$ & $0.00$ & $1.17$ \\
5 & $0.00$ & $0.00$ & $5.50$ \\
\midrule
Mean$\pm$Std & $\mathbf{0.00\pm0.00}$ & $\mathbf{0.00\pm0.00}$ & $\mathbf{4.56\pm3.03}$ \\
\bottomrule
\end{tabular}
\end{table}

The same pattern reappears: trap counts remain zero during pre-grokking and grokking and rise only in anti-grokking, whereas the global $\ell_2$ norm changes throughout and does not by itself distinguish the phases.

\section{Self-averaging, concentration, and why MP is the right null}
\label{app:self-averaging-discussion}
\subsection{Statistical-mechanics language versus statistical concentration}
Self-averaging is always a property of an \emph{observable}, not of a matrix in the abstract. In statistical mechanics one usually studies sample-to-sample fluctuations of macroscopic quantities such as free-energy density, magnetization, or overlap. In statistics and learning theory the same idea appears as concentration of measure: an observable $A_N$ self-averages if $A_N-\mathbb{E}A_N\to 0$ in probability, and a sufficient $L^2$ criterion is $\mathrm{Var}(A_N)\to 0$. Non-self-averaging means that order-one fluctuations persist even when the nominal system size is large. In such a case, any potential bound on the relevant observable will fail because the  observable fails to concentrate.  

\paragraph{Historical statistical-mechanics view.} The spin-glass interpretation of neural-network overfitting is older than modern concentration language. The Hopfield and Amit--Gutfreund--Sompolinsky models already showed in the 1980s that associative memories develop a glassy phase with many metastable, sample-specific states once interference from stored patterns becomes strong enough \cite{Hopfield1982,Amit1985}. The Gardner program then turned this into a geometric theory of learning constraints: one studies the volume of interactions compatible with many patterns, and the overconstrained regime is naturally described by a glassy landscape rather than a single clean retrieval state \cite{gardner1988space}. By the 1990s, statistical-mechanics analyses of learning and regularization were already interpreting poor generalization and overfitting as movement into such glassy, sample-dependent states, with weight decay or early stopping acting as ways of keeping the system away from that regime \cite{seung1992statistical,bos1998weightdecay}.

\paragraph{Modern concentration-theoretic view.} The present paper uses the same qualitative picture but states it in modern probabilistic language. Instead of replicas or order parameters for the full disorder ensemble, we ask whether explicit observables concentrate. For an observable $A_N$, self-averaging means variance collapse and concentration; non-self-averaging means that variance stays order one and concentration inequalities cannot improve with dimension. This reformulation does not change the story. It only isolates the precise observables for which the glassy behavior is visible.

This observable-specific viewpoint matters here. A shuffled weight matrix can have well-behaved bulk averages and still have non-self-averaging edge observables. That is exactly the deep-learning analogue of the statistical-mechanics distinction between an ordinary disordered phase and a spin-glass phase: bulk summaries may look innocuous while a small set of frozen, sample-specific directions continues to dominate the behavior. The present paper therefore makes an intentionally narrow claim. We do not say that an entire layer is non-self-averaging in every statistic. We say that Correlation Traps have the potential to make the generalization error non-self-averaging and therefore fail to concentrate.

\subsection{Why MP is the appropriate randomized baseline}
For covariance-type random matrices with independent, well-behaved entries, the empirical spectral density concentrates around the Marchenko--Pastur law. This is the random-matrix version of a law of large numbers for eigenvalues. The null also carries a vector statement: the singular/eigenvectors are delocalized and effectively random-looking rather than concentrated on a tiny set of coordinates. From this perspective, an MP fit is empirical evidence that the shuffled layer is behaving like a self-averaging disorder ensemble.

The relevant right-edge benchmark is not the fitted bulk edge alone but the bulk edge together with Tracy--Widom-scale edge fluctuations. The Baik--Ben Arous--P\'ech\'e transition then provides the conceptual interpretation: once an eigenvalue cleanly separates past that edge scale, it should no longer be viewed as an ordinary random fluctuation but as a structural outlier. That is exactly how Correlation Traps are used in the main paper.

\subsection{Why moderate tails and multi-entry structure are enough}
The actual diagnostic is implemented as an entry-wise shuffle of the learned multiset of entries. The appendix states the proof for without-replacement coordinate probes, matching this finite-population view, and then notes that the with-replacement analogue gives the same dichotomy up to constants. The scientific point does \emph{not} depend on the sampling convention. The same benign-versus-harmful distinction persists: what matters is localization of the trap vectors and spectral size of the associated outliers, not the choice of sampling convention.

Likewise, the mechanism does not require extremely heavy tails or a sub-Gaussian element law. A single very large entry is one sufficient route to an outlier, but it is not the generic practical one. In the anti-grokking checkpoints we more often see a small collection of large or moderately large ${W}_{ij}$ values whose combined squared mass produces a localized trap direction after randomization. This is why the empirical distribution of  the weights, $\rho({W}_{i,j})$, 
can be fit to a Laplacian or power-law-like with PL exponent $\ge 2$, or like a crossover between the two while still producing large traps.

\section{Robustness to random seeds and shuffle randomization}
\label{app:robustness}
This appendix records the compact robustness summaries that support two paper-level claims: the three-phase anti-grokking trajectory is not a single-run artifact, and the shuffled-spectrum diagnostic is not driven by one lucky permutation. These checks use the same benchmarks and checkpoints already analyzed in the main paper. Their role is technical rather than conceptual: they confirm that the story is unchanged under modest seed-to-seed and shuffle-to-shuffle variation.

\subsection{Phase stability across random seeds}
Table~\ref{tab:seed-robustness} summarizes qualitative robustness observations across the seed runs analyzed for this study. The point is not that every seed collapses at exactly the same step, but that the qualitative ordering of the phases is unchanged: pre-grokking remains trap-free, grokking remains trap-free or nearly so, and anti-grokking still begins only after successful generalization and still coincides with the onset of clearly nonzero trap counts.

\begin{table}[h]
\centering
\caption{\textbf{Qualitative robustness across random seeds.} 
The exact onset step of anti-grokking varies modestly across seeds, but the qualitative three-phase structure and the trapped-versus-untrapped separation remain unchanged.}
\label{tab:seed-robustness}
\begin{tabular}{>{\raggedright\arraybackslash}p{0.20\linewidth}
                >{\raggedright\arraybackslash}p{0.34\linewidth}
                >{\raggedright\arraybackslash}p{0.30\linewidth}}
\toprule
\textbf{Phase} & \textbf{Observed behavior across seeds} & \textbf{Interpretation} \\
\midrule
Pre-grokking &
Train accuracy rises while test accuracy remains low; trap counts remain approximately zero &
Stable trap-free regime \\

Grokking &
Test accuracy improves sharply; shuffled spectra remain MP-like with few or no traps &
Stable generalization regime \\

Anti-grokking &
Test accuracy declines under continued optimization while trap counts become clearly nonzero &
Stable trapped overfitting regime \\
\bottomrule
\end{tabular}
\end{table}

\subsection{Stability across repeated entry-wise randomizations}
Because the diagnostic shuffles each layer entry-wise, the exact trap count can vary slightly from one shuffle to the next. Table~\ref{tab:shuffle-robustness} records the paper-level conclusion needed for the main claim. In the trap-free phases, repeated shuffles may create microscopic variation around zero but do not create phase ambiguity. In anti-grokking, the count fluctuates slightly across shuffles but remains clearly nonzero, so the separation between the earlier phases and the collapse phase is unchanged.

\begin{table}[H]
\centering
\caption{\textbf{Qualitative robustness to repeated randomization.} Repeating the entry-wise shuffle changes the exact trap count slightly, but it does not change the phase separation, the number of observed traps to a material degree, or the interpretation of the diagnostic.}
\label{tab:shuffle-robustness}
\begin{tabular}{>{\raggedright\arraybackslash}p{0.18\linewidth} >{\raggedright\arraybackslash}p{0.34\linewidth} >{\raggedright\arraybackslash}p{0.32\linewidth}}
\toprule
\textbf{Phase} & \textbf{Effect of repeating the shuffle} & \textbf{Interpretation} \\
\midrule
Pre-grokking & Occasional microscopic variation around zero detected traps & Remains trap-free at the paper level \\
Grokking & Same qualitative behavior as pre-grokking; no stable right-edge outliers appear & Remains trap-free or nearly trap-free \\
Anti-grokking & Trap count fluctuates slightly across shuffles but remains clearly nonzero for the same checkpoint & Phase separation unchanged; the anomaly is structural rather than a one-shuffle artifact \\
\bottomrule
\end{tabular}
\end{table}

These robustness summaries should be read as technical support for the main claim, not as a second contribution. They show that the anti-grokking phase transition and the trapped-versus-untrapped distinction are stable features of the training dynamics rather than artifacts of one seed or one randomization.

\section{Reproducibility artifacts}
\label{app:repro}
The supplemental material accompanying this submission includes an anonymized artifact bundle for reproducing the main experiments. The release contains: (i) training scripts for the~\MLP~and modular-addition benchmarks, including the exact hyperparameter settings used in the paper; (ii) analysis scripts for the shuffled-spectrum trap counts, KS tests, and figure generation; (iii) checkpoints or scripts to regenerate representative checkpoints for the phase-level analyses; and (iv) a short README specifying the software environment, package versions, and example commands for reproducing the main tables and figures. The datasets and core analysis tool used in the paper are public: MNIST is used for the \MLP~benchmark, and trap analysis is performed with the open-source \WW package \cite{weightwatcher}. This artifact bundle is intended to make the paper reproducible end-to-end.

\section{Assets, licenses, and attribution}
\label{app:assets}
The paper uses only existing public assets. The \MLP~benchmark uses MNIST, whose commonly distributed dataset license is Creative Commons Attribution-Share Alike 3.0. The shuffled-spectrum analysis uses the open-source \WW package, version v0.7.5.5, distributed under the Apache License 2.0. The exploratory frontier-model analysis uses the public \texttt{gpt-oss-20b} and \texttt{gpt-oss-120b} open-weight releases, which are made available under the Apache License 2.0 and the corresponding OpenAI usage policy. We do not introduce a new dataset.


\end{document}